\pdfoutput=1
\documentclass{article}

\usepackage{microtype}
\usepackage{graphicx}
\usepackage{subfigure}
\usepackage{booktabs} 

\usepackage{hyperref}

\usepackage[accepted]{icml2023}

\usepackage{amsmath}
\usepackage{amssymb}
\usepackage{mathtools}
\usepackage{amsthm}

\usepackage{hyperref}
\usepackage{bbm}
\usepackage{url}
\usepackage{graphicx}
\usepackage{subfigure}
\usepackage{wrapfig}
\usepackage{makecell}
\usepackage{color,array}
\usepackage{bm}
\allowdisplaybreaks[1]

\usepackage[capitalize,noabbrev]{cleveref}

\theoremstyle{plain}

\theoremstyle{definition}

\theoremstyle{remark}

\usepackage[textsize=tiny]{todonotes}

\icmltitlerunning{Multi-task Hierarchical Adversarial Inverse Reinforcement Learning}

\begin{document}

\twocolumn[
\icmltitle{Multi-task Hierarchical Adversarial Inverse Reinforcement Learning}




\begin{icmlauthorlist}
\icmlauthor{Jiayu Chen}{pur}
\icmlauthor{Dipesh Tamboli}{pece}
\icmlauthor{Tian Lan}{gece}
\icmlauthor{Vaneet Aggarwal}{pur,pece,kaust}
\end{icmlauthorlist}

\icmlaffiliation{pur}{School of Industrial Engineering, Purdue University, West Lafayette, IN 47907, USA}
\icmlaffiliation{pece}{Elmore Family School of Electrical and Computer Engineering, Purdue University, West Lafayette, IN 47907, USA}
\icmlaffiliation{gece}{Department of Electrical and Computer Engineering, George Washington University, Washington DC 20052, USA}
\icmlaffiliation{kaust}{Department of Computer Science and AI Initiative, King Abdullah University of Science and Technology, Thuwal 23955, KSA}

\icmlcorrespondingauthor{Jiayu Chen}{chen3686@purdue.edu}

\icmlkeywords{Imitation Learning, Meta Learning, Option}

\vskip 0.3in
]



\printAffiliationsAndNotice{}  

\begin{abstract}
Multi-task Imitation Learning (MIL) aims to train a policy capable of performing a distribution of tasks based on multi-task expert demonstrations, which is essential for general-purpose robots. Existing MIL algorithms suffer from low data efficiency and poor performance on complex long-horizontal tasks. We develop Multi-task Hierarchical Adversarial Inverse Reinforcement Learning (MH-AIRL) to learn hierarchically-structured multi-task policies, which is more beneficial for compositional tasks with long horizons and has higher expert data efficiency through identifying and transferring reusable basic skills across tasks. To realize this, MH-AIRL effectively synthesizes context-based multi-task learning, AIRL (an IL approach), and hierarchical policy learning. Further, MH-AIRL can be adopted to demonstrations without the task or skill annotations (i.e., state-action pairs only) which are more accessible in practice. Theoretical justifications are provided for each module of MH-AIRL, and evaluations on challenging multi-task settings demonstrate superior performance and transferability of the multi-task policies learned with MH-AIRL as compared to SOTA MIL baselines. 
\end{abstract}


 \section{Introduction}

The generalist robot, which can autonomously perform a wide range of tasks, is one of the essential targets of robotic learning. As an important approach, Imitation Learning (IL) enables the agent to learn policies based on expert demonstrations and is especially effective for problems where it's difficult to discover task solutions autonomously through Reinforcement Learning (RL). To train a general-purpose agent, Multi-task/Meta Imitation Learning (MIL) algorithms \cite{DBLP:conf/corl/FinnYZAL17, DBLP:conf/icra/DeisenrothEPF14, DBLP:conf/icra/SinghJIKDLKF20} have been proposed to learn a parameterized policy that is a function of both the current observation and the task and is capable of performing a range of tasks following a particular distribution. The key insight of these algorithms is that the successful control for one task can be informative for other related tasks. However, a critical challenge for them is to acquire enough data for the agent to generalize broadly across tasks. Typically, a large number of demonstrations are required for each task in that distribution, and the required amount increases with task difficulty. Moreover, the learned multi-task policy cannot be transferred to tasks out of that distribution \cite{DBLP:conf/nips/YuYFE19, DBLP:conf/nips/GhasemipourGZ19}, which limits its general use.


Hierarchical Imitation Learning (HIL) has the potential to reduce the required demonstrations. In HIL, the agent learns a two-level policy, which can be modeled with the option framework \cite{DBLP:journals/ai/SuttonPS99}, from the expert data. Specifically, the low-level policies (i.e., skills) are designated to accomplish certain subtasks in a complex task, while the high-level policy is for scheduling the switch among the skills to solve the entire task. For multi-task settings, learning a hierarchical policy enables the agent to identify basic skills that can be useful in solving a distribution of tasks and to transfer them across tasks during training. In this case, each skill can be trained with demonstrations from different tasks rather than limited to a single one, and, with the shared skills, an agent mainly needs to update its high-level policy rather than learning an entire policy for each task. The expert data efficiency is significantly improved since demonstrations among different tasks are reused for learning skills and the burden of multi-task policy learning becomes lower. Further, in RL and IL, hierarchies exhibit a number of benefits, including better performance on long-horizontal complex tasks \cite{DBLP:conf/iclr/FlorensaDA17, DBLP:conf/icml/JingH0MKGL21} and the possibility of skill transfer between distinct tasks \cite{DBLP:conf/icml/AndreasKL17}.



In this paper, we propose MH-AIRL to introduce hierarchies to MIL. As discussed above, such hierarchies can improve expert data efficiency so that the agent can achieve superior performance based on a limited number of demonstrations. Further, basic skills can be extracted from the learned policies and reused in out-of-distribution tasks for better transferability (i.e., addressing the core concern of multi-task learning). For example, it enables locomotion skills to be reused for multiple goal-achieving tasks of the same robot agent, yet in distinct scenarios. Different from previous Multi-task Hierarchical IL (MHIL) algorithms \cite{DBLP:conf/case/FoxBSG19, DBLP:journals/corr/abs-1810-11043, DBLP:conf/corl/GaoJ022, DBLP:conf/iros/BianMH22}, MH-AIRL is context-based and thus can be applied to demonstrations without any (skill or task) annotations, which are more accessible in practice.
To this end, we extend both the multi-task learning and imitation learning modules (i.e., the core components of MIL), with the option framework (i.e., the hierarchical learning module). For multi-task learning, we condition the learned policy on a Hierarchical Latent Context Structure, where the task code and skill segmentation serve as the global and local context variables respectively. To compel the casual relationship of learned policy and latent variables, we start from the definition of mutual information and directed information and derive an easier-to-handle lower bound for each of them, serving as the optimization objectives. For imitation learning, we propose H-AIRL, which redefines a SOTA IL algorithm -- AIRL \cite{DBLP:journals/corr/abs-1710-11248} in an extended state-action space to enable our algorithm to recover a hierarchical policy (rather than a monolithic one) from expert trajectories. Finally, an actor-critic framework -- HPPO is proposed to synthesize the optimization of the three modules above.


The contributions are as follows: (1) Our work presents the first MHIL algorithm based on demonstrations without any (skill or task) annotations, i.e., state-action pairs only. This greatly generalizes the applicability of our algorithm and reduces the cost of building expert datasets. (2) The newly-proposed H-AIRL and HPPO can be independently used for Hierarchical IL and RL, respectively. They are shown to achieve improved performance than SOTA HIL and HRL baselines. (3) We provide theoretical proof and ablation study for each algorithm module, and show the superiority of our algorithm through comparisons with SOTA baselines on a series of challenging multi-task settings from Mujoco \cite{todorov2012mujoco} and D4RL \cite{DBLP:journals/corr/abs-2004-07219}. 



\section{Related Work}

Machine Learning has found successful applications across a wide array of sectors such as transportation \cite{8793143,DBLP:conf/aips/ChenULA21, luo2022multisource, MA2020248}, manufacturing \cite{PEDDIREDDY20211336,fu2021improved}, networking \cite{balachandran2014modeling,geng2023reinforcement}, robotics \cite{DBLP:conf/corl/GaoJ022,gonzalez2023asap}, etc. In the field of robotics, one of the key objectives is developing a `generalist' robot, capable of executing a multitude of tasks with human-like precision. To achieve this, multi-task robotic learning proves to be a highly effective methodology. In this section, we succinctly delineate Multi-task IL and Multi-task HIL, illustrating the contributions and significance of our research in this evolving field.

Multi-task/Meta IL algorithms have been proposed to learn a parameterized policy, which is capable of performing a range of tasks following a particular distribution, from a mixture of expert demonstrations. Based on the meta/multi-task learning techniques used, current MIL algorithms can be categorized as gradient-based or context-based. Gradient-based MIL, such as \cite{DBLP:conf/corl/FinnYZAL17, yu2018one}, integrates a gradient-based meta learning algorithm — MAML \cite{DBLP:conf/icml/FinnAL17} with supervised IL to train a policy that can be fast adapted to a new task with one-step gradient update. Context-based MIL, such as \cite{DBLP:conf/nips/GhasemipourGZ19, DBLP:conf/nips/YuYFE19}, learns a latent variable to represent the task contexts and trains a policy conditioned on the task context variable. Thus, with the corresponding task variable, the policy can be directly adopted to a new task setting. However, these algorithms do not make use of the option framework to learn a hierarchical policy like ours. In Section \ref{ehl}, we compare our algorithm with MIL baselines from both categories and show that it achieves better performance on a wide range of challenging long-horizon tasks. 

Multi-task HIL aims at recovering a multi-task hierarchical policy based on expert demonstrations from a distribution of tasks, which synthesizes the advantages of Multi-task IL and HIL. We present here the previous study in this area. The algorithms proposed in \cite{DBLP:conf/case/FoxBSG19} and \cite{DBLP:journals/corr/abs-2102-09854} are limited to a certain type of robot. They provide predefined subtask decomposition, like picking and placing dishes, to simplify hierarchical learning, and have access to segmented expert demonstrations.  However, our algorithm is proposed to automatically discover a hierarchical policy from unsegmented demonstrations and the discovered policy should capture the subtask structure of the demonstrations without supervision. In \cite{DBLP:journals/corr/abs-1810-11043}, they propose to let the robot learn a series of primitive skills from corresponding demonstrations first, and then learn to compose learned primitives into multi-stage skills to complete a task. Thus, they predefine the types of skills and provide demonstrations corresponding to each skill. Also, in their setting, each new task has to be a sequence of predefined skills. A very recent work \cite{DBLP:conf/corl/GaoJ022} integrates MAML and the option framework for MHIL. Like \cite{DBLP:conf/iros/BianMH22} and \cite{DBLP:conf/nips/DevinGADL19}, this algorithm can be applied to demonstrations without the skill annotations, but these demonstrations have to be categorized by the task, in accordance with the requirements of MAML.  Consequently, our research introduces the first MHIL algorithm that relies on demonstrations devoid of task or skill annotations. This makes it significantly more practical for real-world applications. 

\section{Background}

In this section, we introduce Adversarial Inverse Reinforcement Learning (AIRL), Context-based Meta Learning, and the One-step Option Framework, corresponding to the three components of our algorithm: IL, multi-task learning, and hierarchical policy learning, respectively.  They are based on the Markov Decision Process (MDP), denoted by $\mathcal{M}=(\mathcal{S}, \mathcal{A}, \mathcal{P}, \mu, \mathcal{R}, \mathcal{\gamma})$, where $\mathcal{S}$ is the state space, $\mathcal{A}$ is the action space, $\mathcal{P}:\mathcal{S} \times \mathcal{A} \times \mathcal{S} \rightarrow [0,1]$ is the transition function ($\mathcal{P}^{S_{t+1}}_{S_{t}, A_{t}}\triangleq\mathcal{P}(S_{t+1}|S_{t}, A_{t})$), $\mu:\mathcal{S}\rightarrow [0,1]$ is the distribution of the initial state, $\mathcal{R}:\mathcal{S} \times \mathcal{A} \rightarrow \mathbb{R}$ is the reward function, and $\mathcal{\gamma} \in (0,1]$ is the discount factor. 

\subsection{Adversarial Inverse Reinforcement Learning} \label{airl}

While there are several other ways to perform IL, such as supervised imitation (e.g., Behavioral Cloning (BC) \cite{DBLP:journals/neco/Pomerleau91}) and occupancy matching (e.g., GAIL \cite{DBLP:conf/nips/HoE16}), we adopt Inverse Reinforcement Learning (IRL) because it uses not only the expert data but also self-exploration of the agent with the recovered reward function for further improvement \cite{DBLP:conf/icml/NgR00, DBLP:conf/icra/WangLCLC21}. Comparisons with BC- and GAIL-based algorithms will be provided in Section \ref{eva}. IRL aims to infer an expert's reward function from demonstrations, based on which the expert's policy can be recovered. Maximum Entropy IRL \cite{DBLP:conf/aaai/ZiebartMBD08} solves IRL as a maximum likelihood estimation (MLE) problem shown as Equation \ref{equ:1}. $\tau_{E}\triangleq (S_0, A_0, \cdots, S_T)$ denotes the expert trajectory. $Z_{\vartheta}$ is the partition function which can be calculated with  $Z_{\vartheta}=\sum_{\tau_{E}}\widehat{P}_{\vartheta}(\tau_{E})$.
\begin{equation} \label{equ:1}
\begin{aligned}
        &\mathop{\max}_{\vartheta}\mathbb{E}_{\tau_{E}}\left[\log P_{\vartheta}(\tau_{E})\right]=\mathop{\max}_{\vartheta} \mathbb{E}_{\tau_{E}}\left[\log\frac{\widehat{P}_{\vartheta}(\tau_{E})}{Z_{\vartheta}}\right], \\
         &\ \ \ \ \widehat{P}_{\vartheta}(\tau_{E})=\mu(S_0)\mathop{\prod}_{t=0}^{T-1}\mathcal{P}^{S_{t+1}}_{S_{t}, A_{t}}\exp(\mathcal{R}_{\vartheta}(S_t, A_t))
\end{aligned}
\end{equation}
Since $Z_{\vartheta}$ is intractable for problems with large state-action space, the authors of \cite{DBLP:journals/corr/abs-1710-11248} propose AIRL to solve this MLE problem in a sample-based manner, through alternatively training a discriminator $f_{\vartheta}$ and policy network $\pi$ in an adversarial setting. The discriminator is trained by minimizing the cross-entropy loss between the expert demonstrations $\tau_{E}$ and generated samples $\tau$ by $\pi$:
\begin{equation} \label{equ:3}
\begin{aligned}
    \mathop{\min}_{\vartheta}\mathop{\sum}_{t=0}^{T-1}-\mathbb{E}_{\tau_{E}}\left[\log D_{\vartheta}^t\right]  - \mathbb{E}_{\tau}\left[\log(1-D_{\vartheta}^t)\right]
\end{aligned}
\end{equation}
Here, $D_{\vartheta}^t=D_{\vartheta}(S_t,A_t)=\frac{\exp(f_{\vartheta}(S_t,A_t))}{\exp(f_{\vartheta}(S_t,A_t))+\pi(A_t|S_t)}$. Meanwhile, the policy $\pi$ is trained with RL using the reward function defined as $\log D_{\vartheta}^t-\log(1-D_{\vartheta}^t)$. It is shown that, at optimality, $f_{\vartheta}$ can serve as the recovered reward function $\mathcal{R}_{\vartheta}$ and $\pi$ is the recovered expert policy.

\subsection{Context-based Meta Learning} \label{context}

We consider the Meta IRL setting: given a distribution of tasks $P(\mathcal{T})$, each task sampled from $P(\mathcal{T})$ has a corresponding MDP, and all of them share the same $\mathcal{S}$ and $\mathcal{A}$ but may differ in $\mu$, $\mathcal{P}$, and $\mathcal{R}$. The goal is to train a flexible policy $\pi$ on a set of training tasks sampled from $P(\mathcal{T})$, which can be quickly adapted to unseen test tasks sampled from the same distribution. As a representative, context-based Meta IRL algorithms \cite{DBLP:conf/nips/GhasemipourGZ19, DBLP:conf/nips/YuYFE19} introduce the latent task variable $C$, which provides an abstraction of the corresponding task $\mathcal{T}$,  so each task can be represented with its distinctive components conditioning on $C$, i.e., $(\mu(S_0|C), \mathcal{P}(S'|S,A,C), \mathcal{R}(S,A|C))$. These algorithms learn a context-conditioned policy $\pi(A|S,C)$ from the multi-task expert data, through IRL and by maximizing the mutual information \cite{cover1999elements} between the task variable $C$ and the trajectories from $\pi(A|S,C)$. Thus, given $C$ for a new task, the corresponding $\pi(A|S,C)$ can be directly adopted.
Context-based methods can adopt off-policy data, making them more align with the goal of our work -- learning from demonstrations. Thus, we choose context-based Meta IRL as our base algorithm.


Given expert trajectories sampled from a distribution of tasks (i.e., $C \sim prior(\cdot)$) and assuming that the demonstrative trajectories of each task are from a corresponding expert policy $\pi_{E}(\tau_{E}|C)$, context-based Meta IRL recovers both the task-conditioned reward function $\mathcal{R}_{\vartheta}(S,A|C)$ and policy $\pi(S,A|C)$ by solving an MLE problem:
\begin{equation} \label{equ:6}
\begin{aligned}
        &\ \ \ \ \ \mathop{\max}_{\vartheta}\mathbb{E}_{C\sim prior(\cdot),\tau_{E} \sim \pi_{E}(\cdot|C)}\left[\log P_{\vartheta}(\tau_{E}|C)\right],\\
    &P_{\vartheta}(\tau_{E}|C)\propto \mu(S_0|C)\prod_{t=0}^{T-1}\mathcal{P}^{S_{t+1}}_{S_{t}, A_{t}, C}\ e^{\mathcal{R}_{\vartheta}(S_t, A_t|C)}
\end{aligned}
\end{equation}
where $\mathcal{P}^{S_{t+1}}_{S_{t}, A_{t}, C}\triangleq\mathcal{P}(S_{t+1}|S_{t}, A_{t}, C)$.
Like Equation \ref{equ:1}, this can be efficiently solved through AIRL. We provide the AIRL framework to solve Equation \ref{equ:6} in Appendix \ref{c-airl}.

\subsection{One-step Option Framework} \label{option}

As proposed in \cite{DBLP:journals/ai/SuttonPS99}, an option $Z \in \mathcal{Z}$ can be described with three components: an initiation set $I_Z \subseteq \mathcal{S}$, an intra-option policy $\pi_Z(A|S): \mathcal{S} \times \mathcal{A} \rightarrow [0,1]$, and a termination function $ \beta_Z(S): \mathcal{S} \rightarrow [0,1]$. An option $Z$ is available in state $S$ if and only if $S \in I_Z$. Once the option is taken, actions are selected according to $\pi_Z$ until it terminates stochastically according to $\beta_Z$, i.e., the termination probability at the current state. A new option will be activated by a high-level policy $\pi_\mathcal{Z}(Z|S): \mathcal{S} \times \mathcal{Z} \rightarrow [0,1]$ once the previous option terminates. In this way, $\pi_\mathcal{Z}(Z|S)$ and $\pi_Z(A|S)$ constitute a hierarchical policy for a certain task. Hierarchical policies tend to have superior performance on complex long-horizontal tasks which can be broken down into a series of subtasks \cite{chen2022odpp, 9847387,NEURIPS2022_c40d1e40, DBLP:journals/corr/abs-2210-03269}.

The one-step option framework \cite{li2020skill} is proposed to learn the hierarchical policy without the extra need to justify the exact beginning and breaking condition of each option, i.e., $I_Z$ and $\beta_Z$. First, it assumes that each option is available at each state, i.e., $I_Z=\mathcal{S}, \forall Z \in \mathcal{Z}$. Second, it drops $\beta_Z$ through redefining the high-level and low-level (i.e., intra-option) policies as $\pi_{\theta}(Z|S, Z')$ ($Z'$: the option in the last timestep) and $\pi_{\phi}(A|S,Z)$ respectively and implementing them as end-to-end neural networks with the Multi-Head Attention (MHA) mechanism \cite{DBLP:conf/nips/VaswaniSPUJGKP17}, which enables it to temporally extend options in the absence of the termination function. Intuitively, if $Z'$ still fits $S$, $\pi_{\theta}(Z|S, Z')$ will assign a larger attention weight to $Z'$ and thus has a tendency to continue with it; otherwise, a new option with better compatibility will be sampled. Then, the option is sampled at each timestep rather than after the last one terminates. With this simplified framework, we only need to train the hierarchical policy, i.e., $\pi_{\theta}$ and $\pi_{\phi}$, of which the structure design with MHA is in Appendix \ref{mha}.

\section{Proposed Approach}

In this section, we propose Multi-task Hierarchical AIRL (MH-AIRL) to learn a multi-task hierarchical policy from a mixture of expert demonstrations. \textbf{First}, the learned policy is multi-task by conditioning on the task context variable $C$. Given $C \sim prior(\cdot)$, the policy can be directly adopted to complete the corresponding task. In practice, we can usually model a class of tasks by specifying the key parameters of the system and their distributions (i.e., $prior(C)$), including the property of the agent (e.g., mass and size), circumstance (e.g., friction and layout), and task setting (e.g., location of the goals). In this case, directly recovering a policy, which is applicable to a class of tasks, is quite meaningful. \textbf{Second}, for complex long-horizontal tasks which usually contain subtasks, learning a monolithic policy to represent a structured activity can be challenging and inevitably requires more demonstrations. In contrast, a hierarchical policy can make full use of the subtask structure and has the potential for better performance. Moreover, the learned low-level policies can be transferred as basic skills to out-of-distribution tasks for better transferability, while the monolithic policy learned with previous Meta IL algorithms cannot.

In Section \ref{info} and \ref{h-airl}, we extend context-based Meta Learning and AIRL with the option framework, respectively. In Section \ref{frame}, we synthesize the three algorithm modules and propose an actor-critic framework for optimization.

\subsection{Hierarchical Latent Context Structure} \label{info}

\begin{figure}
\centering
\includegraphics[width=3.0in, height=1.2in]{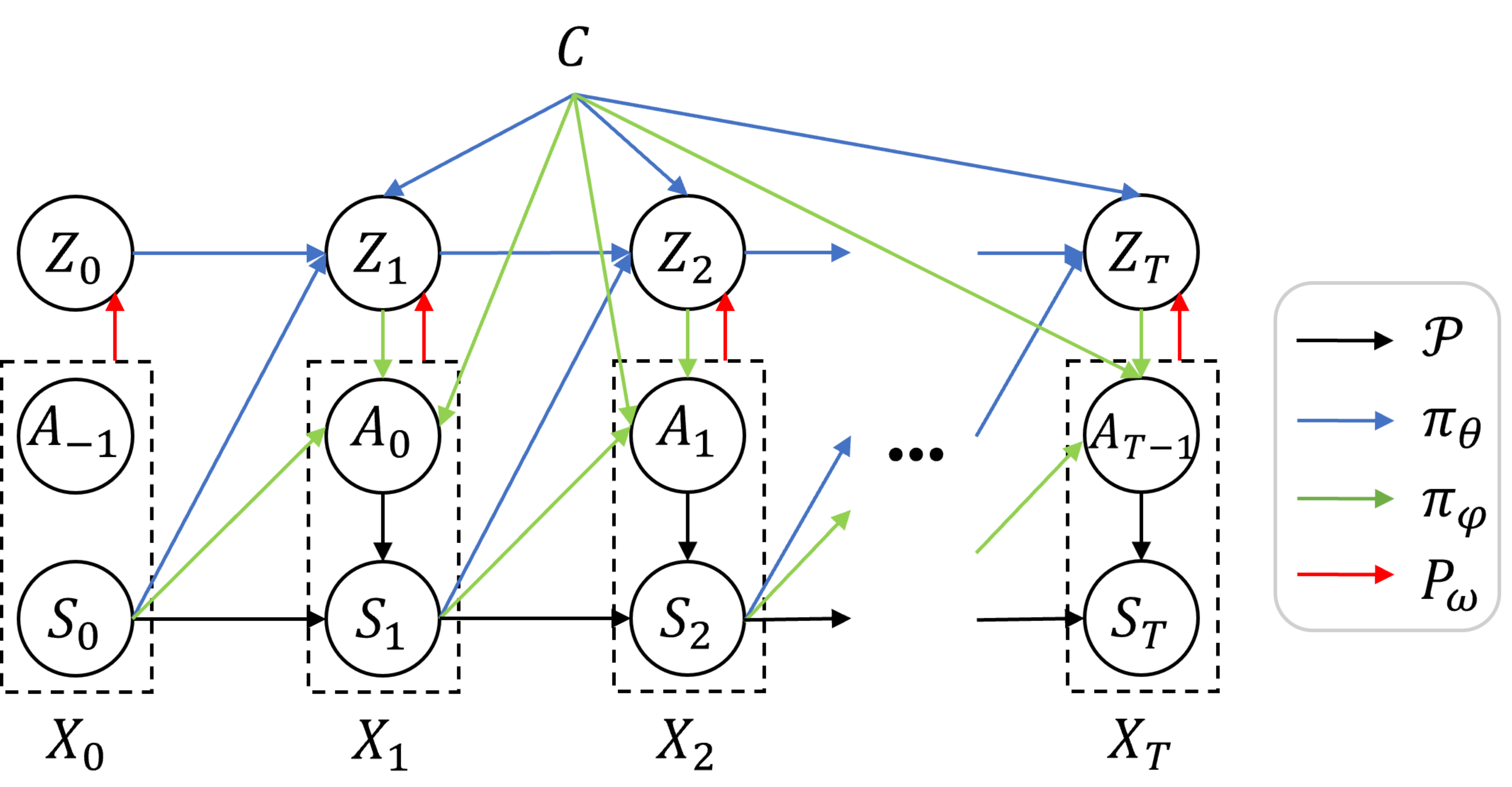}
\caption{Illustration of the hierarchical latent context structure and its implementation with the one-step option model.}
\label{fig:1}
\end{figure}

As mentioned in Section \ref{context}, the current task for the agent is encoded with the task variable $C$, which serves as the \textbf{global} context since it is consistent through the episode. As mentioned in Section \ref{option}, at each step, the hierarchical policy agent will first decide on its option choice $Z$ using $\pi_{\theta}$ and then select the primitive action based on the low-level policy $\pi_{\phi}$ corresponding to $Z$. In this case, the policy learned should be additionally conditioned on $Z$ besides the task code $C$, and the option choice is specific to each timestep $t \in \{0,\cdots,T\}$, so we view the option choices $Z_{0:T}$ as the \textbf{local} latent contexts. $C$ and $Z_{0:T}$ constitute a hierarchical latent context structure shown as Figure \ref{fig:1}. 
Moreover, real-world tasks are often compositional, so the agent requires to reason about the subtask at hand while dealing with the global task. $Z_{0:T}$ and $C$ provide a hierarchical embedding, which enhances the expressiveness of the policy trained with MH-AIRL, compared with context-based Meta IL which only employs the task context. \textbf{In this section}, we define the mutual and directed information objectives to enhance the causal relationship between the hierarchical policy and the global \& local context variables which the policy should condition on, as an extension of context-based Meta-IL with the one-step option model.


Context-based Meta IL algorithms establish a connection between the policy and task variable $C$, so that the policy can be adapted among different task modes according to the task context. This can be realized through maximizing the mutual information between the trajectory generated by the policy and the corresponding $C$, i.e., $I(X_{0:T};C)$, where $X_{0:T} = (X_0, \cdots, X_T) = ((A_{-1}, S_0), \cdots, (A_{T-1},S_T))=\tau$. $A_{-1}$ is a dummy variable.  On the other hand, the local latent variables $Z_{0:T}$ have a directed causal relationship with the trajectory $X_{0:T}$ shown as the probabilistic graphical model in Figure \ref{fig:1}. As discussed in \cite{massey1990causality, DBLP:conf/iclr/SharmaSRK19}, this kind of connection can be established by maximizing the directed information (a.k.a., causal information) flow from the trajectory to the latent factors of variation, i.e., $I(X_{0:T} \rightarrow Z_{0:T})$. In our multi-task framework, we maximize the conditional directed information $I(X_{0:T} \rightarrow Z_{0:T}|C)$, since for each task $c$, the corresponding $I(X_{0:T} \rightarrow Z_{0:T}|C=c)$ should be maximized. 

Directly optimizing the mutual or directed information objective is computationally infeasible, so we instead maximize their variational lower bounds as follows: (Please refer to Appendix \ref{dilb} and \ref{milb} for the definition of mutual and directed information and derivations of their lower bounds. For simplicity, we use $X^T$ to represent $X_{0:T}$, and so on.)
\begin{equation} \label{equ:14}
\begin{aligned}
        &L^{MI} \triangleq H(C)+ \mathop{\mathbb{E}}_{X^T, Z^T, C}\log P_{\psi}(C|X_{0:T}) \\
        &L^{DI} \triangleq \sum_{t=1}^{T} [\mathop{\mathbb{E}}_{X^t, Z^t, C} \log P_{\omega}(Z_t|X_{0:t}, Z_{0:t-1}, C) \\
        &\qquad\qquad\qquad + H(Z_t|X_{0:t-1}, Z_{0:t-1}, C)] 
\end{aligned}
\end{equation}
where $H(\cdot)$ denotes the entropy, $P_{\psi}$ and $P_{\omega}$ are the variational estimation of the posteriors $P(C|X_{0:T})$ and $P(Z_t|X_{0:t}, Z_{0:t-1}, C)$ which cannot be calculated directly.
$P_{\psi}$ and $P_{\omega}$ are implemented as neural networks, $H(C)$ is constant, and $H(Z_t|X_{0:t-1}, Z_{0:t-1}, C)$ is the entropy of the output of the high-level policy network (Appendix \ref{dilb}), so $L^{MI}$ and $L^{DI}$ can be computed in real-time.
Moreover, the expectation on $X^t, Z^t, C$ in $L^{MI}$ and $L^{DI}$ can be estimated in a Monte-Carlo manner \cite{sutton2018reinforcement}: $C \sim prior(\cdot)$, $(X_{0:t}, Z_{0:t}) \sim P_{\theta, \phi}(\cdot|C)$, where $P_{\theta, \phi}(X_{0:t}, Z_{0:t}|C)$ is calculated by: (See Appendix \ref{dilb}.)
\begin{equation} \label{equ:16}
\begin{aligned}
    \setlength{\abovedisplayskip}{0.5pt}
    \setlength{\belowdisplayskip}{0.5pt}
    &\mu(S_{0}|C)\mathop{\prod}_{i=1}^{t}[\pi_{\theta}(Z_{i}|S_{i-1}, Z_{i-1}, C) \cdot\\
    &\pi_{\phi}( A_{i-1}|S_{i-1}, Z_{i}, C)\mathcal{P}^{S_i}_{S_{i-1},A_{i-1}, C}]
\end{aligned}
\end{equation}

Combining Equation \ref{equ:14} and \ref{equ:16}, we can get the objectives with respect to $\pi_{\theta}$ and $\pi_{\phi}$, i.e., the hierarchical policy defined in the one-step option model. By maximizing $L^{MI}$ and $L^{DI}$, the connection between the policy and the hierarchical context structure can be established and enhanced. In $L^{MI}$ and $L^{DI}$, we also introduce two variational posteriors $P_{\psi}$ and $P_{\omega}$ and update them together with $\pi_{\theta}$ and $\pi_{\phi}$. An analogy of our learning framework with Variational Autoencoder (VAE) \cite{DBLP:journals/corr/KingmaW13} is provided in Appendix \ref{rnn-imp}, which provides another perspective to understand the proposed objectives.

\subsection{Hierarchical AIRL} \label{h-airl}

In this section, we consider how to recover the task-conditioned hierarchical policy from a mixture of expert demonstrations $\{(X_{0:T}^E, Z_{0:T}^E, C_E)\}$. Current algorithms, like AIRL \cite{DBLP:journals/corr/abs-1710-11248} or Meta AIRL \cite{DBLP:conf/nips/GhasemipourGZ19, DBLP:conf/nips/YuYFE19}, can not be directly adopted since they don't take the local latent codes $Z_{0:T}^E$ into consideration. Thus, we propose a novel hierarchical extension of AIRL, denoted as H-AIRL, as a solution, which is also part of our contributions. Further, it's usually difficult to annotate the local and global latent codes, i.e., $Z_{0:T}^E$ and $C_E$, of an expert trajectory $X_{0:T}^E$, so we propose an Expectation-Maximization (EM) adaption of H-AIRL as well to learn the multi-task hierarchical policy based on only the unstructured expert trajectories $\{X_{0:T}^E\}$.

First, we define the task-conditioned hierarchical policy. When observing a state $S_t$ at timestep $t \in \{0,\cdots, T-1\}$ during a certain task $C$, the agent needs first to decide on its option choice based on $S_t$ and its previous option choice $Z_t$ using the high-level policy $\pi_{\theta}(Z_{t+1}|S_t,Z_t,C)$, and then decide on the action with the corresponding low-level policy $\pi_{\phi}(A_t|S_t,Z_{t+1},C)$. Thus, the task-conditioned hierarchical policy can be acquired with the chain rule as:
\begin{equation} \label{equ:7}
\begin{aligned}
    &\ \ \ \ \ \pi_{\theta}(Z_{t+1}|S_t,Z_t,C)\cdot\pi_{\phi}(A_t|S_t,Z_{t+1},C)\\
    &=\pi_{\theta, \phi}(Z_{t+1},A_{t}|S_{t}, Z_{t}, C)=\pi_{\theta, \phi}(\widetilde{A}_t|\widetilde{S}_t, C)
\end{aligned}
\end{equation}
where the first equality holds because of the one-step Markov assumption (i.e., $\pi_{\phi}(A_t|S_t,Z_{t},Z_{t+1},C)=\pi_{\phi}(A_t|S_t,Z_{t+1},C)$), $\widetilde{S}_t \triangleq (S_t,Z_{t})$ and $\widetilde{A}_t \triangleq (Z_{t+1},A_{t})$ denote the extended state and action space respectively.

Next, by substituting $(S_t,A_t)$ with $(\widetilde{S}_t, \widetilde{A}_t)$ and $\tau_E$ with the hierarchical trajectory $(X_{0:T}, Z_{0:T})$ in Equation \ref{equ:6}, we can get an MLE problem shown as Equation \ref{equ:8}, from which we can recover the task-conditioned hierarchical reward function and policy. The derivation is in Appendix \ref{mle-obj}.
\begin{equation} \label{equ:8}
\begin{aligned}
        &\mathop{\max}_{\vartheta}\mathbb{E}_{C, (X^T, Z^T) \sim \pi_{E}(\cdot|C)}\left[\log P_{\vartheta}(X^T, Z^T|C)\right], \\
    &\quad\  P_{\vartheta}(X_{0:T}, Z_{0:T}|C)\propto \widehat{P}_{\vartheta}(X_{0:T}, Z_{0:T}|C) \\
&=\mu(S_0|C)\mathop{\prod}_{t=0}^{T-1}\mathcal{P}^{S_{t+1}}_{S_t, A_{t}, C}\ e^{\mathcal{R}_{\vartheta}(S_t, Z_{t}, Z_{t+1}, A_{t}|C)}
\end{aligned}
\end{equation}
Equation \ref{equ:8} can be efficiently solved with the adversarial learning framework shown as Equation \ref{equ:10} ($C, C_E\sim prior(\cdot), (X_{0:T}^E, Z_{0:T}^E) \sim \pi_{E}(\cdot|C_E)$, and $(X_{0:T}, Z_{0:T}) \sim \pi_{\theta, \phi}(\cdot|C)$). At optimality, we can recover the hierarchical policy of the expert as $\pi_{\theta, \phi}$ with these objectives, of which the justification is provided in Appendix \ref{just}.
\begin{equation} \label{equ:10}
\begin{aligned}
       & \mathop{\min}_{\vartheta}-\mathbb{E}_{C_E, (X_{0:T}^E, Z_{0:T}^E)}\mathop{\sum}_{t=0}^{T-1}\log D_{\vartheta}(\widetilde{S}_t^E,\widetilde{A}_t^E|C_E) \\
        &\ \ \ \ \ \ \ \ - \mathbb{E}_{C, (X_{0:T}, Z_{0:T})}\mathop{\sum}_{t=0}^{T-1}\log(1-D_{\vartheta}(\widetilde{S}_t,\widetilde{A}_t|C)), \\
        &\qquad\quad \mathop{\max}_{\theta, \phi}L^{IL}=\mathbb{E}_{C, (X_{0:T}, Z_{0:T})}\mathop{\sum}_{t=0}^{T-1}R_{IL}^{t}
\end{aligned}
\end{equation}
where the reward function $R_{IL}^{t}=\log D_{\vartheta}^t - \log(1-D_{\vartheta}^t)$ and $D_{\vartheta}^t = D_{\vartheta}(\widetilde{S}_t,\widetilde{A}_t|C)=\frac{\exp(f_{\vartheta}(\widetilde{S}_t,\widetilde{A}_t|C))}{\exp(f_{\vartheta}(\widetilde{S}_t,\widetilde{A}_t|C))+\pi_{\theta, \phi}(\widetilde{A}_t|\widetilde{S}_t,C)}$.

In practice, the unstructured expert data $\{X_{0:T}^E\}$, i.e., trajectories only, is more accessible. In this case, we can view the latent contexts as hidden variables in a hidden Markov model (HMM) \cite{EDDY1996361} shown as Figure \ref{fig:1} and adopt an EM-style adaption to our algorithm, where we use the variational posteriors introduced in Section \ref{info} to sample the corresponding $C_E, Z_{0:T}^E$ for each $X_{0:T}^E$. In the E step, we sample the global and local latent codes with $C_E \sim P_{\overline{\psi}}(\cdot|X_{0:T}^E), Z_{0:T}^E \sim P_{\overline{\omega}}(\cdot|X_{0:T}^E,C_E)$. $P_{\overline{\psi}}$ and $P_{\overline{\omega}}$ represent the posterior networks for $C$ and $Z_{0:T}$ respectively, with the parameters $\overline{\psi}$ and $\overline{\omega}$, i.e., the old parameters before being updated in the M step. Then, in the M step, we optimize the hierarchical policy and posteriors with Equation \ref{equ:14} and \ref{equ:10}. Note that the expert data used in the first term of Equation \ref{equ:10} should be replaced with $(X_{0:T}^E, Z_{0:T}^E, C_E)$ collected in the E step. By this adaption, we can get the solution of the original MLE problem (Equation \ref{equ:8}), i.e., the recovered expert policy $\pi_{\theta, \phi}$, with only unstructured expert data, which is proved in Appendix \ref{em-airl}.

\subsection{Overall Framework} \label{frame}

In Section \ref{info}, we propose $L^{MI}(\theta, \phi, \psi)$ and $L^{DI}(\theta, \phi, \omega)$ to establish the causal connection between the policy and hierarchical latent contexts. Then, in Section \ref{h-airl}, we propose H-AIRL to recover the hierarchical policy from multi-task expert demonstrations, where the policy is trained with the objective $L^{IL}(\theta,\phi)$. In this section, we introduce our method to update the hierarchical policy and posteriors with these objectives, and describe the overall algorithm framework. Detailed derivations of $\nabla_{\theta,\phi,\psi}L^{MI}$, $\nabla_{\theta,\phi,\omega}L^{DI}$ and $\nabla_{\theta,\phi}L^{IL}$ are in Appendix \ref{mi-gra}, \ref{di-gra}, and \ref{il-gra}, respectively.

\begin{figure*}[t]
\centering
\subfigure[Mujoco Env]{
\label{fig:4(a)} 
\includegraphics[width=1.8in, height=1.0in]{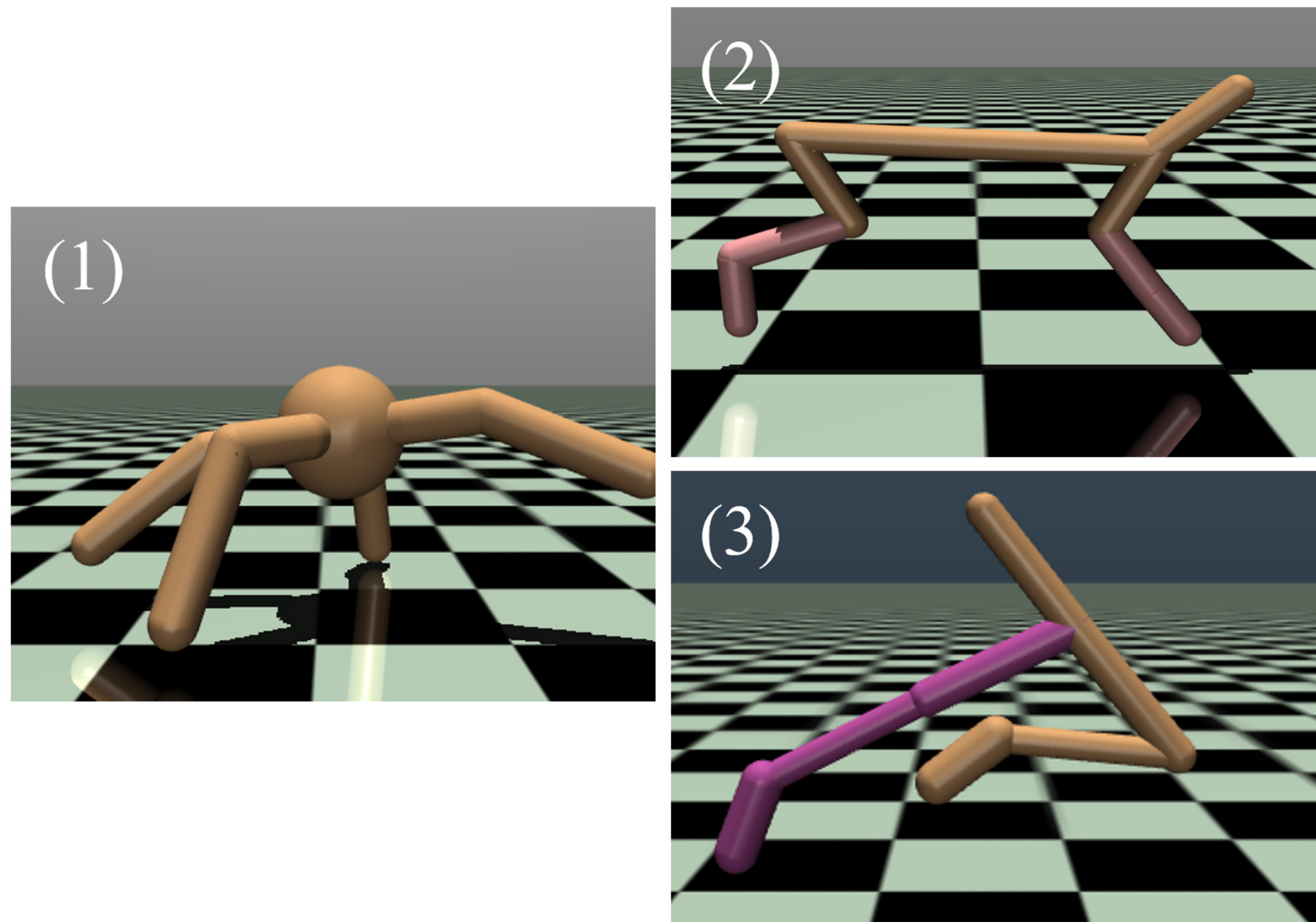}}
\subfigure[HalfCheetah-MultiVel]{
\label{fig:4(b)} 
\includegraphics[width=2.3in, height=1.0in]{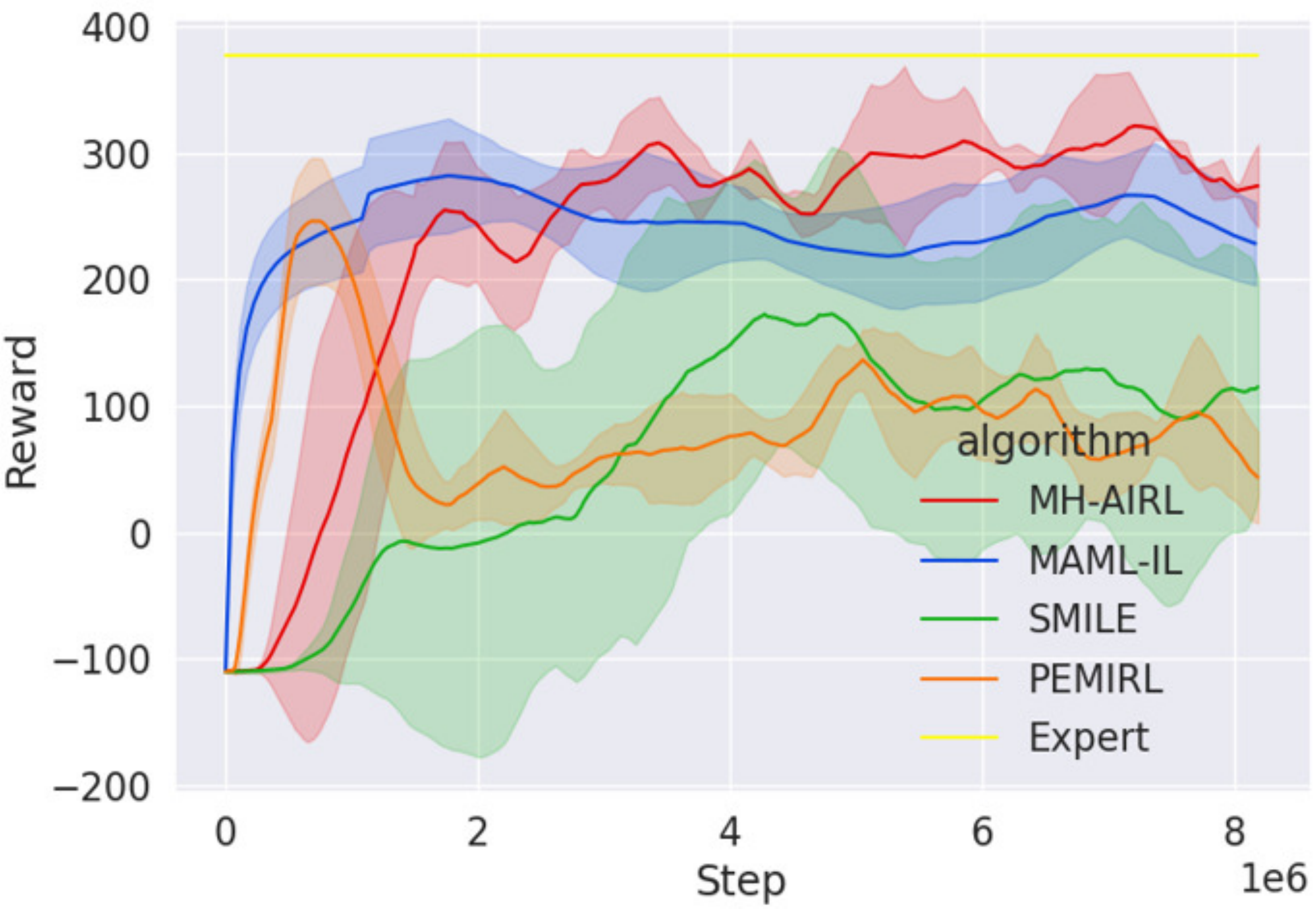}}
\subfigure[Walker-RandParam]{
\label{fig:4(c)} 
\includegraphics[width=2.3in, height=1.0in]{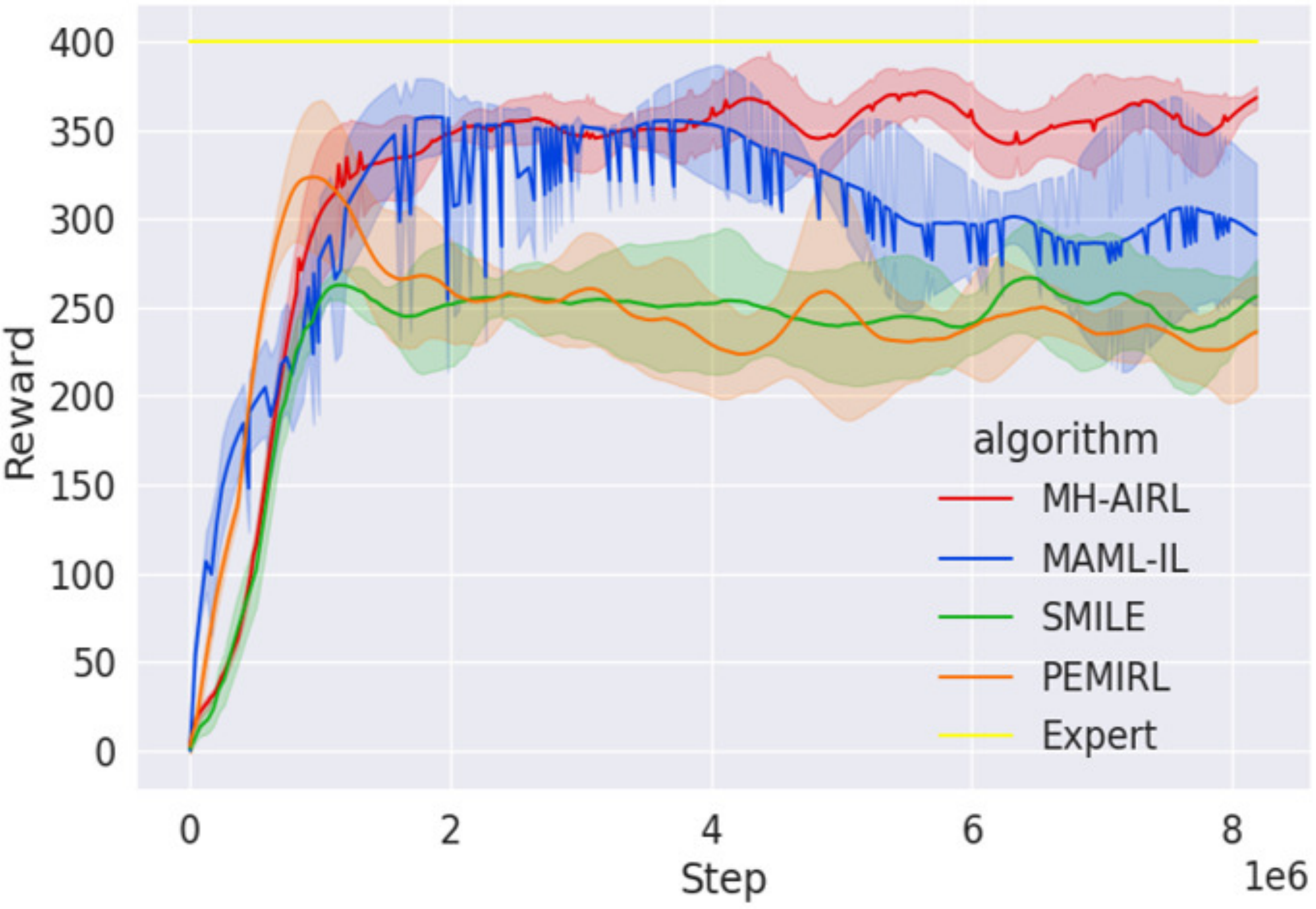}}
\subfigure[Kitchen Env]{
\label{fig:4(d)} 
\includegraphics[width=1.8in, height=1.0in]{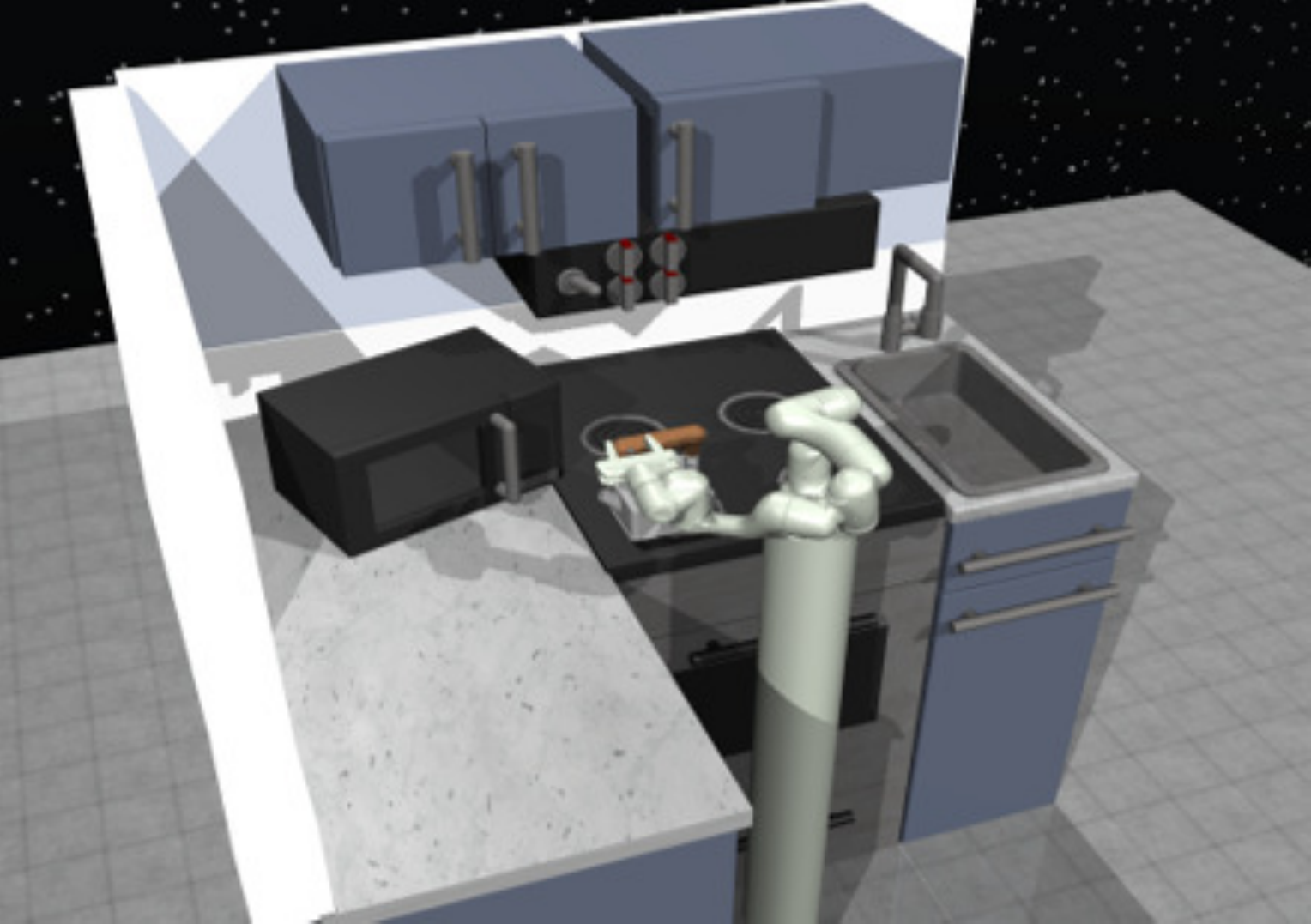}}
\subfigure[Ant-MultiGoal]{
\label{fig:4(e)} 
\includegraphics[width=2.3in, height=1.0in]{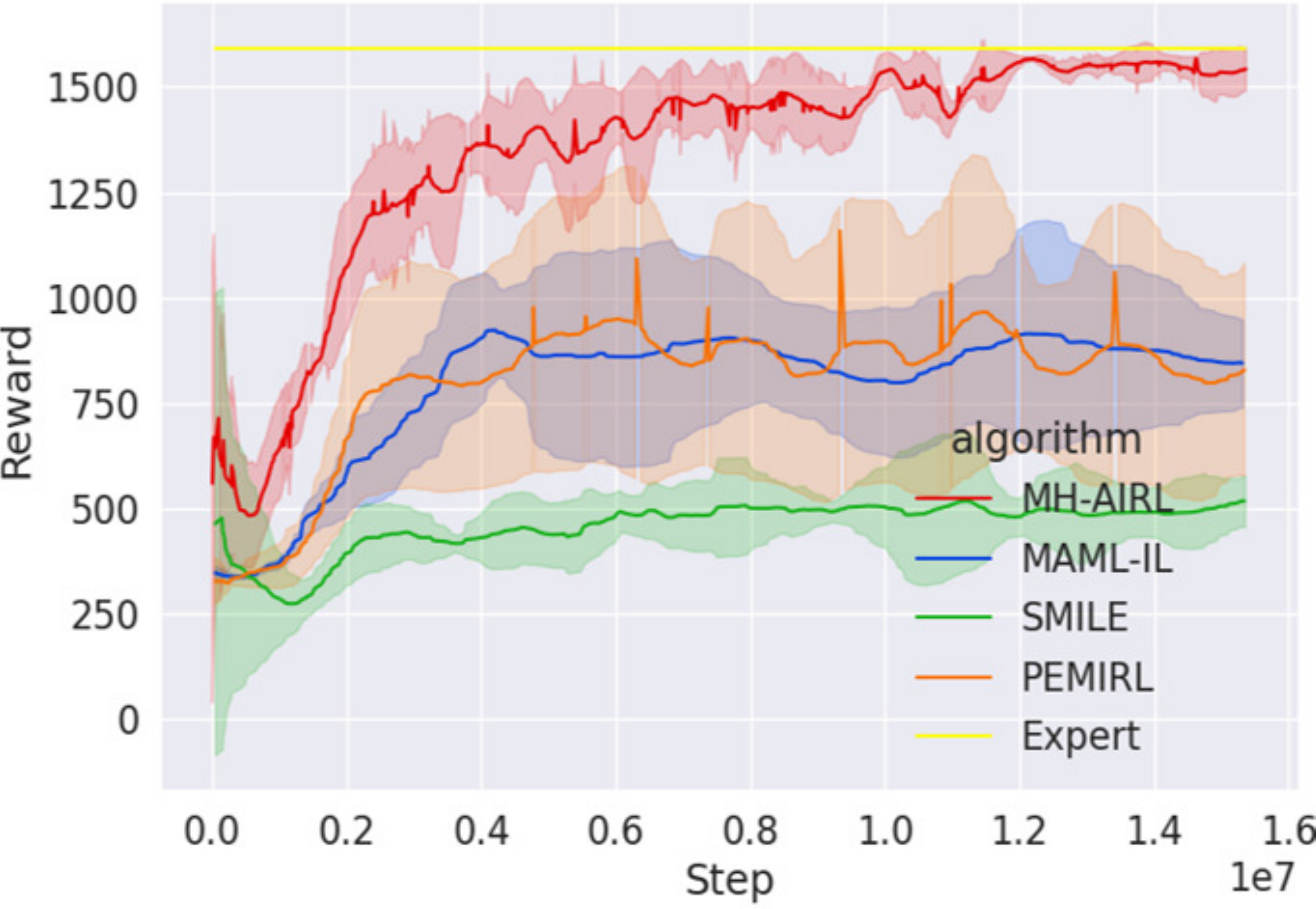}}
\subfigure[Kitchen-MultiSeq]{
\label{fig:4(f)} 
\includegraphics[width=2.3in, height=1.0in]{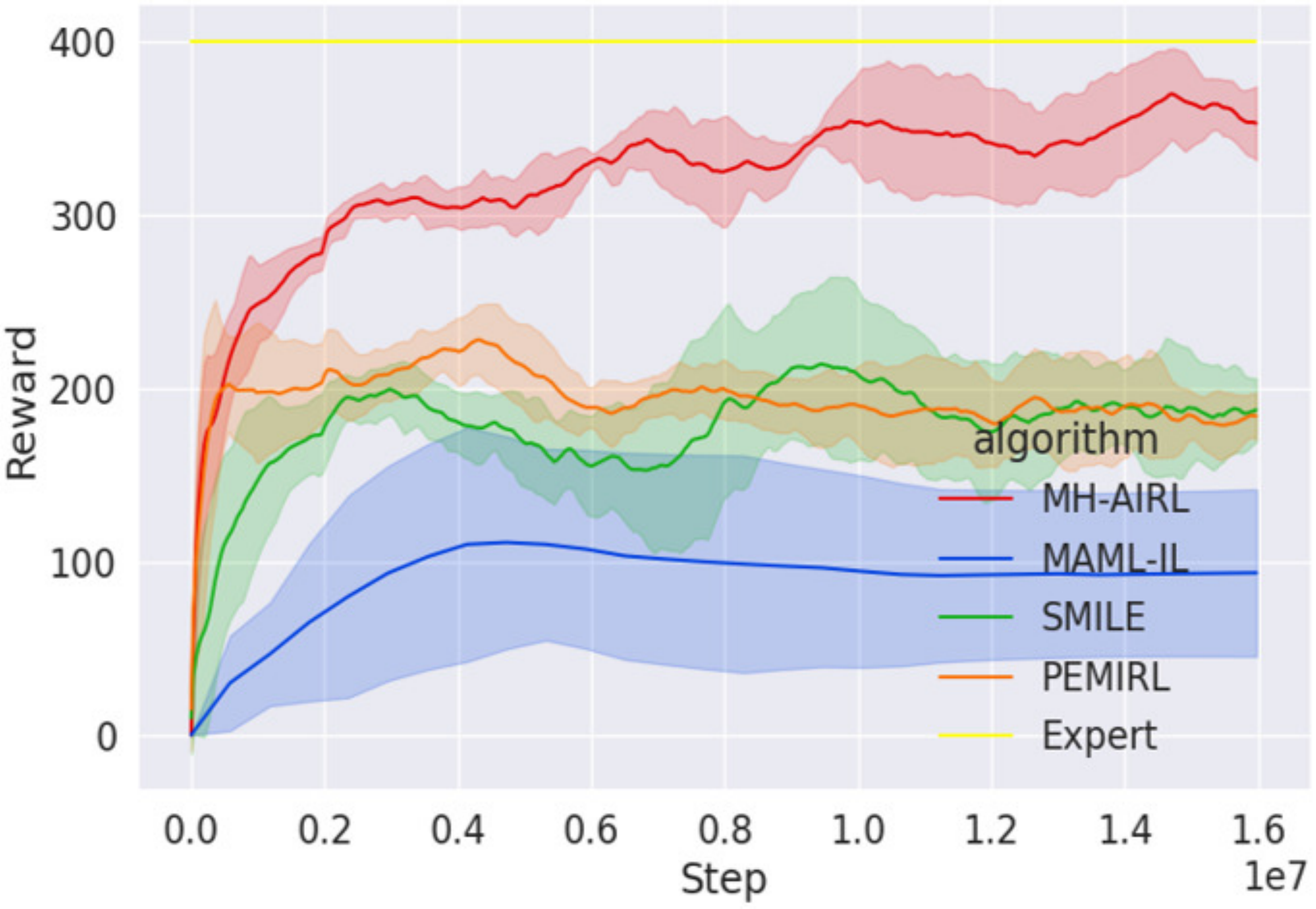}}
\caption{(a) Multi-stage Mujoco locomotion tasks, where (1)-(3) show Ant, HalfCheetah, and Walker agent, respectively. (d) The Kitchen task. (b)(c)(e)(f) Comparison results of MH-AIRL with SOTA Meta Imitation Learning baselines on the four challenging tasks.}
\label{fig:4} 
\end{figure*}

First, the variational posteriors $P_{\psi}$ and $P_{\omega}$ can be updated with the gradients shown in Equation \ref{equ:17} through Stochastic Gradient Descent (SGD) \cite{bottou2010large}.
\begin{equation} \label{equ:17}  
\begin{aligned}
    &\qquad \nabla_{\psi}L^{MI}=\mathop{\mathbb{E}}_{\substack{C,X^T,Z^T}}\nabla_{\psi}\log P_{\psi}(C|X_{0:T})\\
    &\nabla_{\omega}L^{DI}=\mathop{\sum}_{t=1}^{T} \mathop{\mathbb{E}}_{\substack{C,X^t,Z^t}}\nabla_{\omega}\log P_{\omega}(Z_t|X^{t}, Z^{t-1}, C)
\end{aligned}
\end{equation}
Next, the gradients with respect to $\theta$ and $\phi$, i.e., the hierarchical policy, are computed based on the overall objective: 
\begin{equation} \label{equ:19}  
\begin{aligned}
    L=\alpha_1 L^{MI} + \alpha_2 L^{DI} + \alpha_3 L^{IL}
\end{aligned}
\end{equation}
where $\alpha_{1:3}$ are the weights (only the ratios $\frac{\alpha_1}{\alpha_3}, \frac{\alpha_2}{\alpha_3}$ matter) and fine-tuned as hyperparameters.  Based on $L$, we can get the unbiased gradient estimators with respect to $\theta$ and $\phi$: (Derivations are in Appendix \ref{overall}.)
    \begin{equation} \label{equ:a25}  
\begin{aligned}
    &\nabla_{\theta}L = \mathop{\mathbb{E}}_{\substack{C,X^T,Z^T}}[\mathop{\sum}_{t=1}^{T}\nabla_{\theta}\log\pi_{\theta}(Z_{t}|S_{t-1}, Z_{t-1}, C) \cdot \\
    &\qquad\qquad\qquad\quad(Ret_t-b^{high}(S_{t-1}, Z_{t-1}|C))]\\
    &\nabla_{\phi}L = \mathop{\mathbb{E}}_{\substack{C,X^T, Z^T}}[\mathop{\sum}_{t=1}^{T}\nabla_{\phi}\log\pi_{\phi}( A_{t-1}|S_{t-1}, Z_{t}, C) \cdot \\
   &\qquad\qquad\qquad\quad(Ret_t-b^{low}(S_{t-1}, Z_{t}| C))]\\
\end{aligned}
\end{equation}
\begin{equation} \label{equ:a250}  
\begin{aligned}
   &Ret_t = \alpha_1 \log P_{\psi}(C|X_{0:T})\\
   &+\mathop{\sum}_{i=t}^{T}[\alpha_2 \log\frac{P_{\omega}(Z_i|X^{i}, Z^{i-1}, C)}{ \pi_{\theta}(Z_i|S_{i-1},Z_{i-1},C)}+\alpha_3 R_{IL}^{i-1}]
\end{aligned}
\end{equation}
$Ret_t$ represents the return at timestep $t$, while $b^{high}$ and $b^{low}$ are the baseline terms for training $\pi_{\theta}$ and $\pi_{\phi}$, respectively. Further, we claim that the advantage functions for training $\pi_{\theta}$ and $\pi_{\phi}$ are given by $Ret_t - b^{high}(S_{t-1}, Z_{t-1}|C)$ and $Ret_t-b^{low}(S_{t-1}, Z_{t}| C)$, respectively, based on which we can optimize the hierarchical policy via off-the-shelf RL algorithms. In our implementation, we adopt PPO \cite{DBLP:journals/corr/SchulmanWDRK17} to train $\pi_{\theta}$ and $\pi_{\phi}$ with their corresponding advantage functions, respectively. This forms a novel Hierarchical RL (HRL) algorithm -- HPPO, which has shown superiority over RL and HRL baselines in our experiment.

In Appendix \ref{illu}, we provide the overall algorithm as Algorithm \ref{alg:1} and illustrate the interactions among the networks in MH-AIRL in Figure \ref{fig:3}.


\section{Evaluation and Main Results} \label{eva}

MH-AIRL is proposed to learn a multi-task hierarchical policy from a mixture of (unstructured) expert demonstrations. The learned policy can be adopted to any task sampled from a distribution of tasks. In this section: (1) We provide an ablation study with respect to the three main components of our algorithm: context-based multi-task/meta learning, option/hierarchical learning, and imitation learning. (2) We show that the hierarchical policy learning can significantly improve the agent's performance on challenging long-horizontal tasks. (3) Through qualitative and quantitative results, we show that our algorithm can capture the subtask structure within the expert demonstrations and that the learned basic skills for the subtasks (i.e., options) can be transferred to tasks not within the task distribution to aid learning, for better transferability.

The evaluation is based on three Mujoco \cite{todorov2012mujoco} locomotion tasks and the Kitchen task from the D4RL benchmark \cite{DBLP:journals/corr/abs-2004-07219}. All of them are with continuous state \& action spaces, and contain compositional subtask structures to make them long-horizontal and a lot more challenging. To be specific: (1) In \textbf{HalfCheetah-MultiVel}, the goal velocity $v$ is controlled by a 1-dim Gaussian context variable. The HalfCheetah agent is required to speed up to $v/2$ first, then slow down to 0, and finally achieve $v$. (2) In \textbf{Walker-RandParam}, the Walker agent must achieve the goal velocity 4 in three stages, i.e., [2, 0, 4]. Meanwhile, the mass of the agent changes among different tasks, which is controlled by an 8-dim Gaussian context variable. (3) In \textbf{Ant-MultiGoal}, a 3D Ant agent needs to reach a certain goal, which is different in each task and controlled by a 2-dim Gaussian context variable (polar coordinates). Moreover, the agent must go through certain subgoals. For example, if the goal is $(x, y)$ and $|x|>|y|$, the agent must go along $[(0, 0), (x, 0), (x, y)]$. (4) In \textbf{Kitchen-MultiSeq}, there are seven different subtasks, like manipulating the microwave, kettle, cabinet, switch, burner, etc. Each task requires the sequential completion of four specific subtasks. Twenty-four permutations are chosen and so 24 tasks, each of which is sampled with the same probability and controlled by a discrete context variable (input as one-hot vectors). Note that the states of the robot agents only contain their original states (defined by Mujoco or D4RL) and the task context variable, and do not include the actual task information, like the goal (velocity) and subgoal list. The task information is randomly generated by a parametric model of which the parameter is used as the context variable (i.e., the Gaussian vectors as mentioned above). The mapping between context variables and true task information is unknown to the learning agent. This makes the learning problem more challenging and our algorithm more general, since a vector of standard normal variables can be used to encode multiple types of task information.

These scenarios are designed to evaluate our algorithm on a wide range of multi-task setups. First, the agent needs to adapt across different reward functions in (1) and (3) since the rewarding state changes, and adjust across different transition functions in (2) since the mass change will influence the robotic dynamics. Next, different from (1)-(3), discrete context variables are adopted in (4), and (4) provides more realistic and challenging robotic tasks for evaluation. The expert data for Mujoco tasks are from expert agents trained with an HRL algorithm \cite{DBLP:conf/nips/ZhangW19a} and specifically-designed rewards. While for the Kitchen task, we use the human demonstrations provided by \cite{DBLP:conf/corl/0004KLLH19}. Note that the demonstrations (state-action pairs only) do not include the rewards, task or option variables. Codes for reproducing all the results are on \href{https://github.com/LucasCJYSDL/Multi-task-Hierarchical-AIRL}{https://github.com/LucasCJYSDL/Multi-task-Hierarchical-AIRL}.


\begin{table*}[t]
\centering
\caption{Numeric results of the ablation study}
\begin{tabular}{c || c c c c}
\hline
{ } & {HalfCheetah-MultiVel} & {Walker-RandParam} & {Ant-MultiGoal} & {Kitchen-MultiSeq}\\
\hline
\hline
{Expert} & {$376.55 \pm 11.12$} & {$399.95 \pm 1.43$} & {$1593.17 \pm 40.91$} & {$400.00 \pm 0.00$}\\
\hline
{MH-AIRL (ours)} & {\bm{$292.79 \pm 15.99$}} & {\bm{$357.59 \pm 12.10$}} & {\bm{$1530.82 \pm 15.18$}} & {\bm{$352.59 \pm 15.12$}}\\
\hline
\makecell[c]{MH-GAIL (ours)}
 & {$211.32 \pm 52.74$} & {$268.92 \pm 49.29$} & {$1064.78 \pm 180.28$} & {$212.13 \pm 25.25$} \\  
\hline
\makecell[c]{H-AIRL (ours)}
 & {$126.85 \pm 21.92$} & {$225.48 \pm 12.87$} & {$533.80 \pm 40.69$} & {$83.97 \pm 10.95$} \\  
\hline
\makecell[c]{Option-GAIL}
 & {$-44.89 \pm 51.95$} & {$132.01 \pm 54.75$} & {$383.05 \pm 13.52$} & {$204.73 \pm 56.41$} \\  
\hline
\makecell[c]{DI-GAIL}
 & {$56.77 \pm 49.76$} & {$225.22 \pm 14.01$} & {$328.06 \pm 19.89$} & {$131.79 \pm 53.29$} \\  
\hline
\end{tabular}
\label{table:1}
\end{table*}

\subsection{Effect of Hierarchical Learning} \label{ehl}

In this part, we evaluate whether the use of options can significantly improve the learning for challenging compound multi-task settings. We compare MH-AIRL with SOTA Meta Imitation Learning (MIL) baselines which also aim to train a policy that can be fast adapted to a class of related tasks but does not adopt options in learning. Context-based MIL, such as \textbf{PEMIRL} \cite{DBLP:conf/nips/YuYFE19} and \textbf{SMILE} \cite{DBLP:conf/nips/GhasemipourGZ19}, learns a context-conditioned policy that can be adopted to any task from a class by applying the task variable. While the policy learned with Gradient-based MIL, such as \textbf{MAML-IL} \cite{DBLP:conf/corl/FinnYZAL17} which integrates MAML \cite{DBLP:conf/icml/FinnAL17} (a commonly-adopted Meta Learning algorithm) and Behavioral Cloning (BC), has to be updated with gradients calculated from trajectories of the new task, before being applied. We select PEMIRL, SMILE, and MAML-IL from the two major categories of MIL as our baselines. All the algorithms are trained with the same expert data, and evaluated on the same set of test tasks (not contained in the demonstrations). Note that, unlike the others, MAML-IL requires expert data of each test task besides the task variable when testing and requires the expert demonstrations to be categorized by the task when training, which may limit its use in practical scenarios. Our algorithm is trained based on unstructured demonstrations and is only provided with the task context variable for testing. 

In Figure \ref{fig:4}, we record the change of the episodic reward (i.e., the sum of rewards for each step in an episode) on the test tasks as the number of training samples increases. The training is repeated 5 times with different random seeds for each algorithm, of which the mean and standard deviation are shown as the solid line and shadow area, respectively. Our algorithm outperforms the baselines in all tasks, and the improvement is more significant as the task difficulty goes up (i.e., in Ant \& Kitchen), which shows the effectiveness of hierarchical policy learning especially in complex tasks. MAML-IL makes use of more expert information in both training and testing, but its performance gets worse on more challenging tasks. This may be because it is based on BC, which is a supervised learning algorithm prone to compounding errors \cite{DBLP:journals/jmlr/RossGB11}.


\subsection{Ablation Study} \label{abs}
We proceed to show the effectiveness of the IL and context-based multi-task learning components through an ablation study. We propose two ablated versions of our algorithm: (1) \textbf{MH-GAIL} -- a variant by replacing the AIRL component of MH-AIRL with GAIL \cite{DBLP:conf/nips/HoE16} (another commonly-used IL algorithm), of which the details are in Appendix \ref{mh-gail}. (2) \textbf{H-AIRL} -- a version that does not consider the task context $C$, which means $P_{\psi}$ (i.e., the posterior for $C$) is not adopted, $L^{MI}$ is eliminated from Equation \ref{equ:19}, and other networks do not use $C$ as input. H-AIRL can be viewed as a newly-proposed HIL algorithm since it integrates the option framework and IL. To be more convincing, we also use two SOTA HIL algorithms -- \textbf{Option-GAIL} \cite{DBLP:conf/icml/JingH0MKGL21} and \textbf{DI-GAIL} \cite{DBLP:conf/iclr/SharmaSRK19}, as the baselines. The training with the HIL algorithms is based on the same multi-task expert data as ours.


In Appendix \ref{ab-plots}, we provide the plots of the change of episodic rewards on the test tasks. The training with each algorithm is repeated for 5 times with different random seeds. For each algorithm, we compute the average episodic reward after the learning converges in each of the 5 runs, and record the mean and standard deviation in Table \ref{table:1} as the convergence performance. First, we can see that our algorithm performs the best on all tasks over the ablations, showing the effectiveness of all the main modules of our algorithm. Second, MH-GAIL performs better than HIL baselines, showing the necessity of including the context-based multi-task learning component. Without this component, HIL algorithms can only learn an average policy for a class of tasks from the mixture of multi-task demonstrations. Last, H-AIRL, the newly-proposed HIL algorithm, performs better than the SOTA HIL baselines on Mujoco tasks. A comprehensive empirical study on H-AIRL is provided in \cite{chen2022hierarchical}.

\begin{figure*}[t]
\centering
\subfigure[PointCell-MultiGoal]{
\label{fig:6(a)} 
\includegraphics[width=1.8in, height=1.0in]{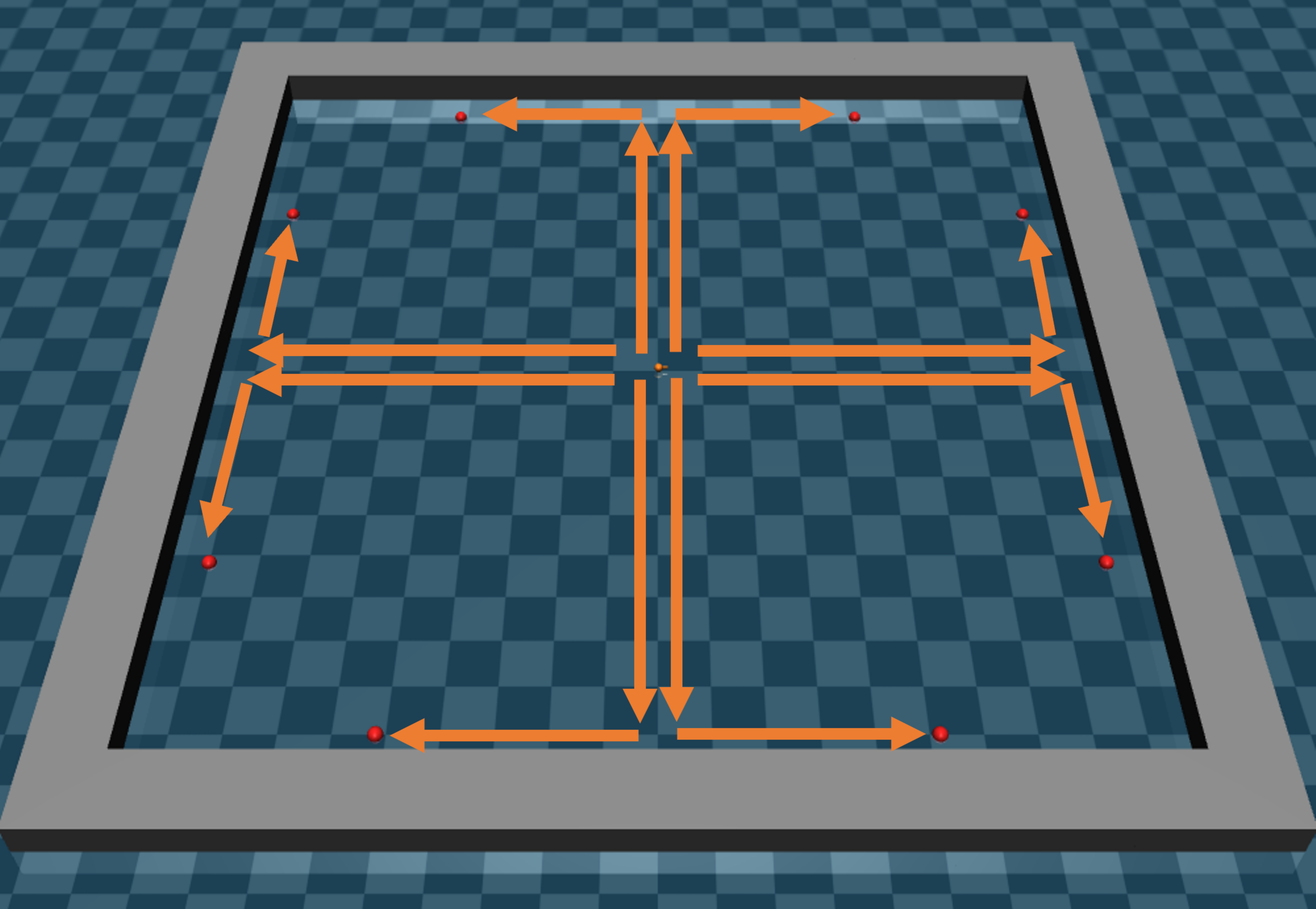}}
\subfigure[PointRoom]{
\label{fig:6(b)} 
\includegraphics[width=1.8in, height=1.0in]{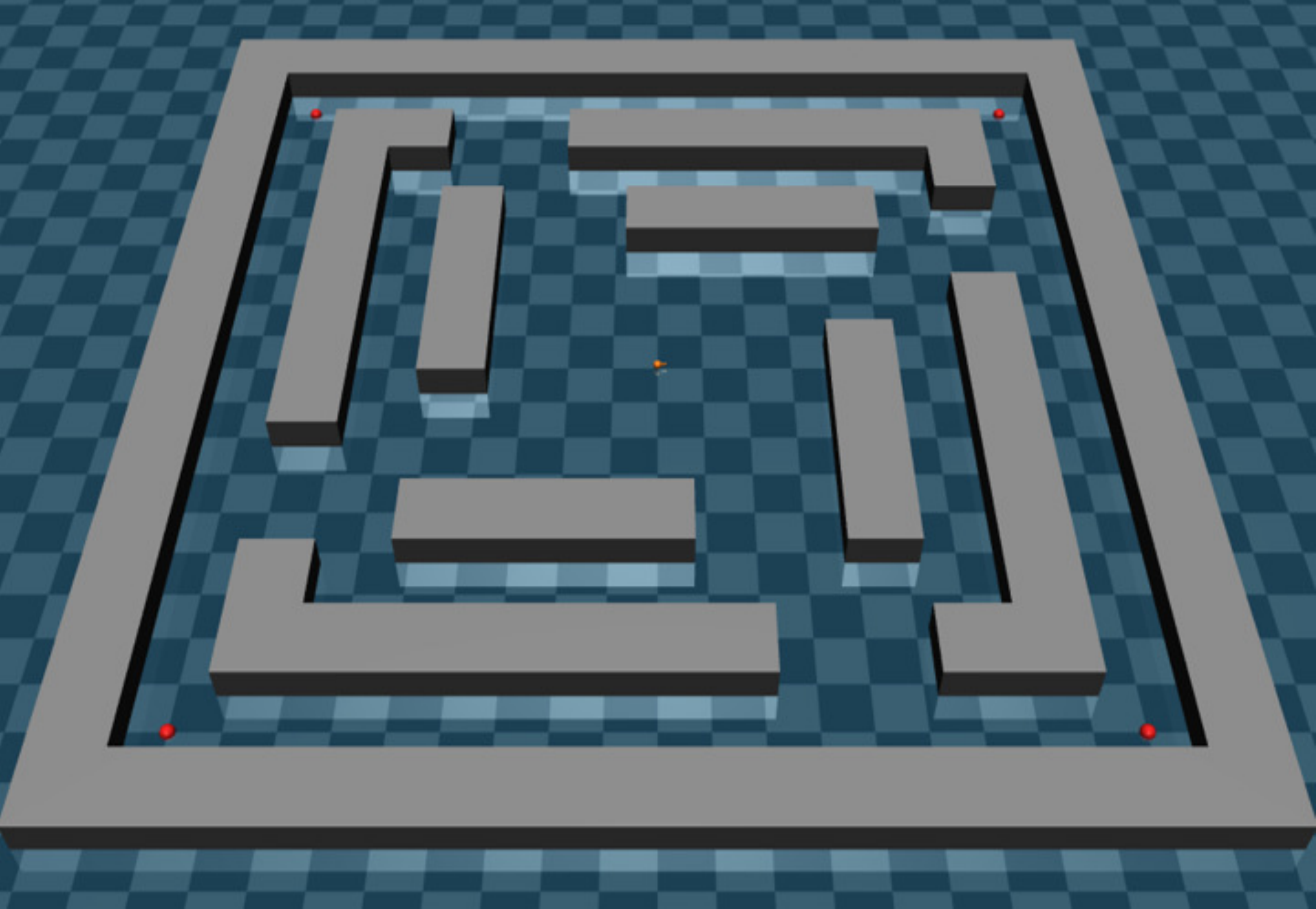}}
\subfigure[PointMaze]{
\label{fig:6(c)} 
\includegraphics[width=1.8in, height=1.0in]{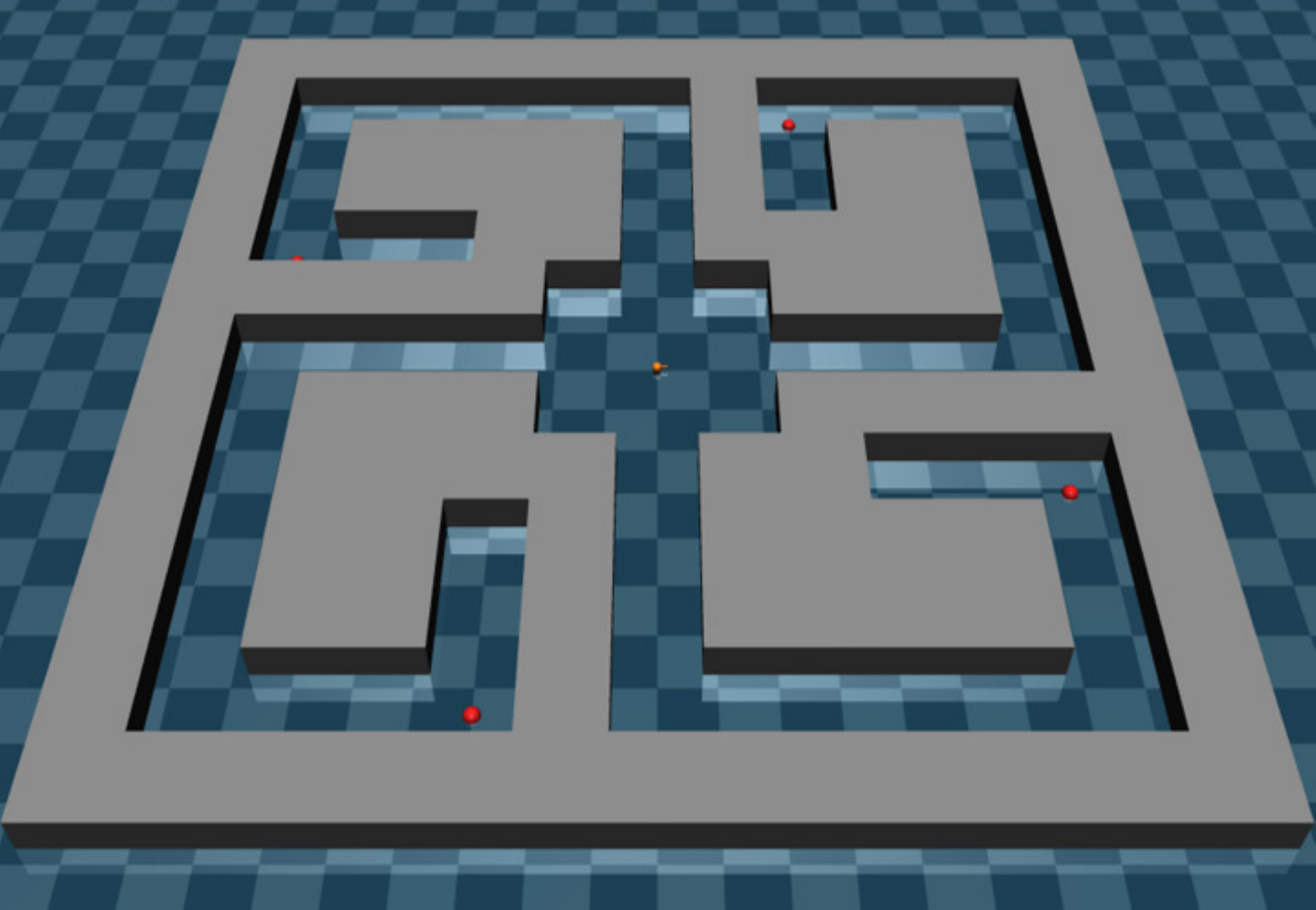}}
\subfigure[Recovered Hier Policy]{
\label{fig:6(d)} 
\includegraphics[width=1.4in, height=1.0in]{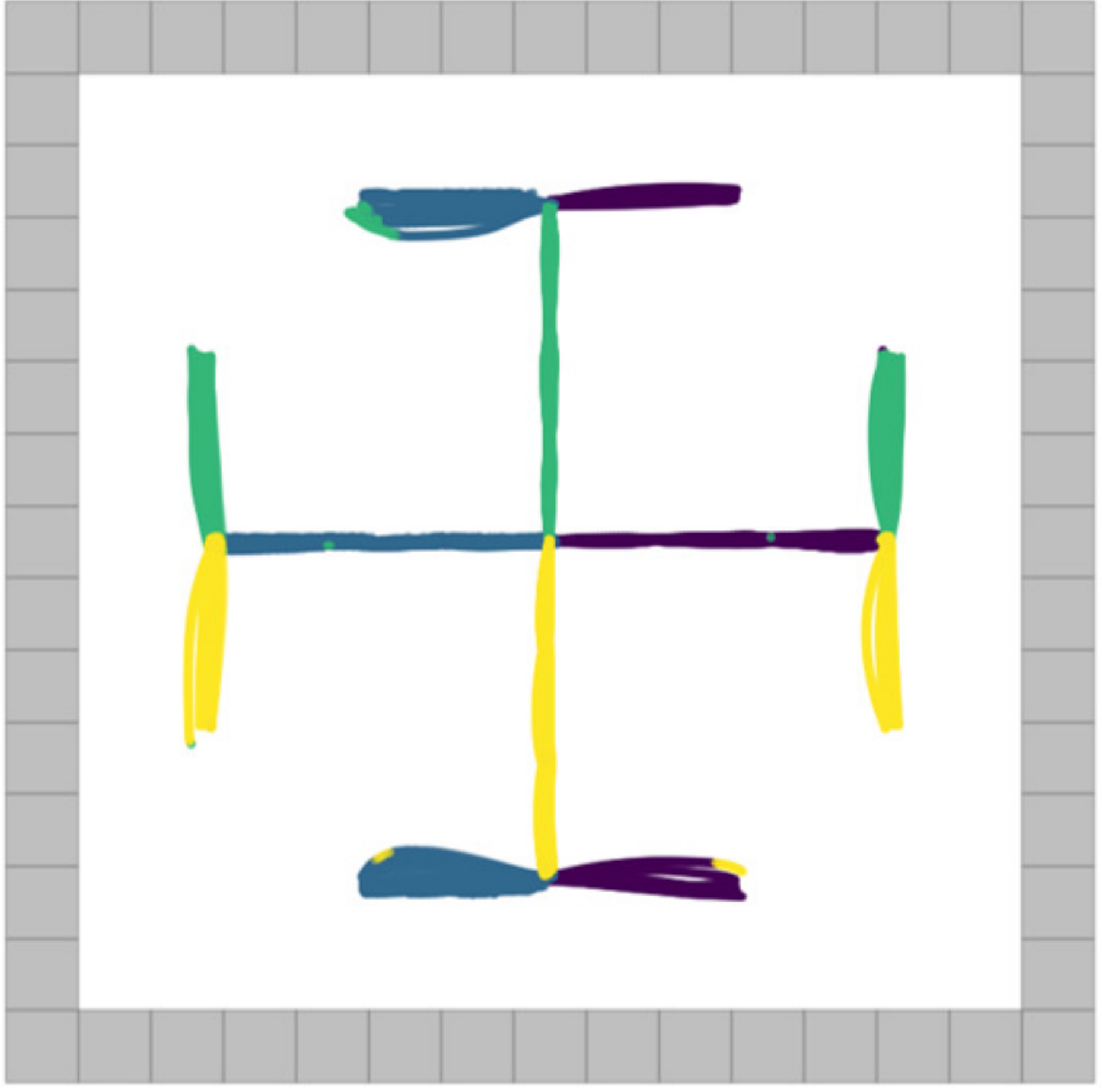}}
\subfigure[Transfer Learning on PointRoom]{
\label{fig:6(e)} 
\includegraphics[width=2.0in, height=1.0in]{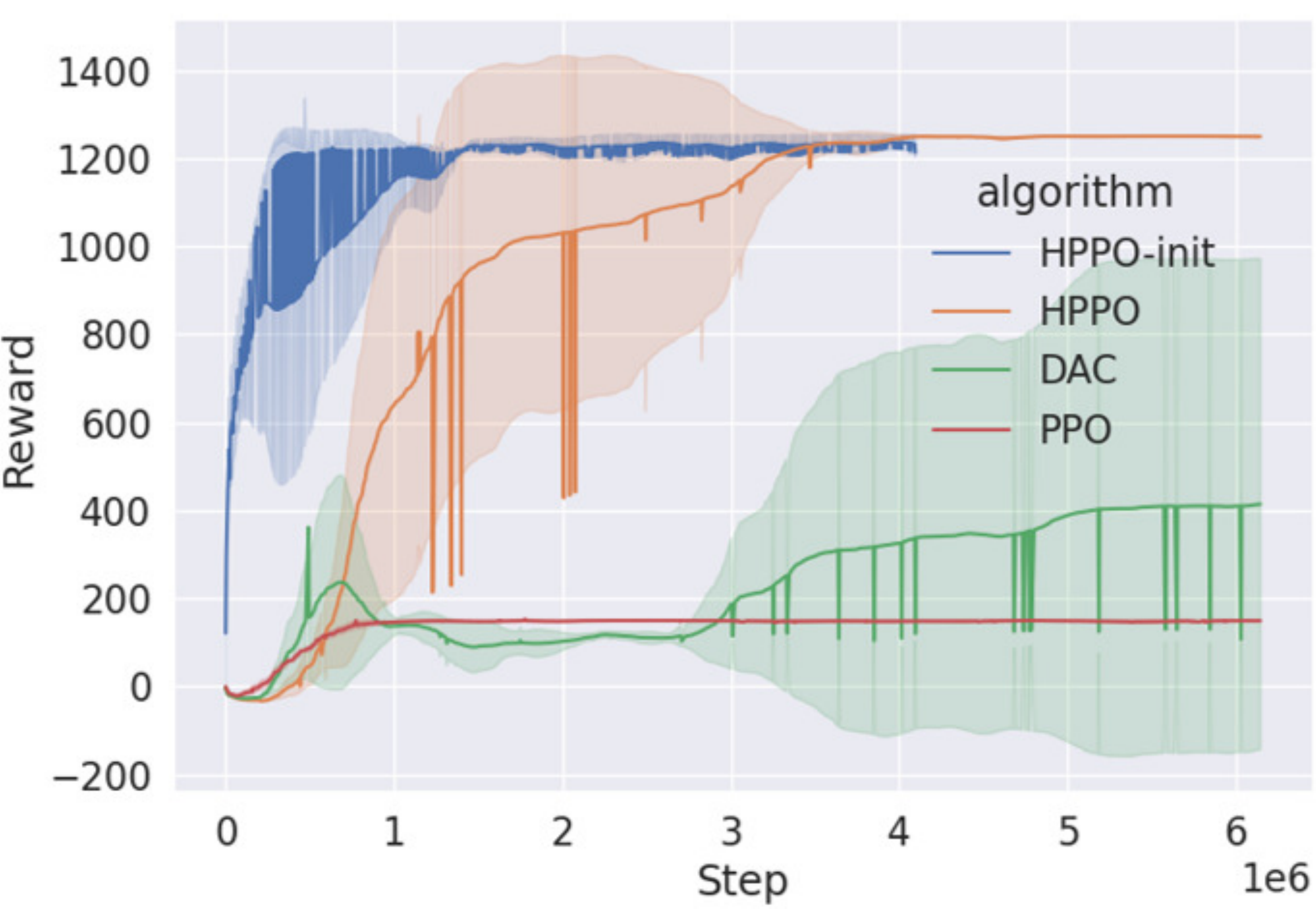}}
\subfigure[Transfer Learning on PointMaze]{
\label{fig:6(f)} 
\includegraphics[width=2.0in, height=1.0in]{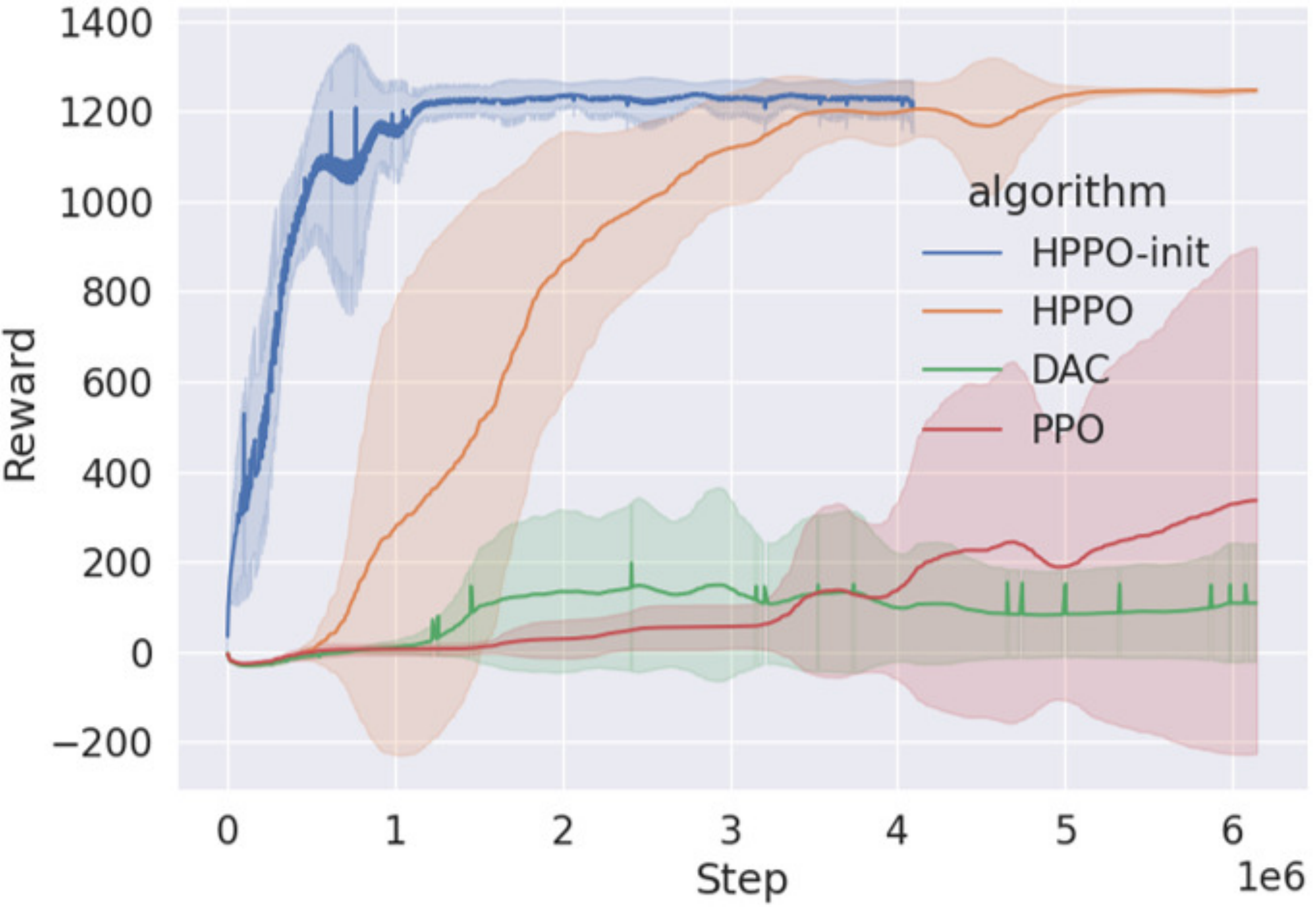}}
\caption{(a) The environment for multi-task learning with MH-AIRL. (d) Visualization of the learned hierarchical policy in  (a). (b)(c) New task settings for evaluating the learned options in (a). (e)(f) Comparison results on (b)(c) between our proposed HRL algorithm (i.e., HPPO) initialized with the transferred options (i.e., HPPO-init) and other SOTA HRL and RL baselines.}
\label{fig:6} 
\end{figure*}

\subsection{Analysis on the Learned Hierarchical Policy} \label{alhp}
In this section, we do the case study to analyze if the learned hierarchical policy can capture the sub-task structure in the demonstrations, and if the learned options can be transferred to tasks out of the task distribution. Capturing the subtask structures in real-life tasks can be essential for the (multi-task) policy learning, because: (1) It is more human-like to split a complex task into more manageable subtasks to learn separately and then synthesize these skills to complete the whole task. (2) In some circumstances, the basic skills learned from one task setting can be reused in other task settings so the agent only needs to update its high-level policy over the same skill set, significantly lowering the learning difficulty. We test our algorithm on Mujoco-MultiGoal (Figure \ref{fig:6(a)}) where the agent is required to achieve a goal corresponding to the task variable (2-dim Gaussian). The expert demonstrations include 100 goal locations in the Cell and the expert agent only moves horizontally or vertically. We test the learned hierarchical policy on 8 sparsely distributed goal locations, of which the trajectories are shown as Figure \ref{fig:6(d)}. We can see: (1) Four options (labeled with different colors) are discovered based on the demonstrations, each of which corresponds to a particular forward direction (green: up, yellow: down, etc.). These options are shared among the tasks. (2) The agent knows how to switch among the options to complete the tasks in stages (i.e., horizontal and vertical) with the learned high-level policy. Thus, our algorithm can effectively capture the compositional structure within the tasks and leverage it in the multi-task policy learning, which explains its superior performance. 
More analysis results of the learned hierarchical policy on HalfCheetah-MultiVel and Walker-RandParam are in Appendix \ref{halfwalker}.

Next, previous Meta/Multi-task Learning algorithms can learn a policy for a class of tasks whose contexts follow a certain distribution, but the learned policy cannot be transferred as a whole to tasks out of this class. In contrast, our algorithm recovers a hierarchical policy, of which the low-level part can be reused as basic skills for new tasks not necessarily in the same class, resulting in substantially improved transferability of the learned policy. To show this, we reuse the options discovered in PointCell as the initialization of the low-level part of the hierarchical policy for the goal-achieving tasks in new scenarios -- PointRoom and PointMaze (Figure \ref{fig:6(b)} and \ref{fig:6(c)}). In each scenario, we select 4 challenging goals (starting from the center point) for evaluation, which are labeled as red points in the figure. Unlike the other evaluation tasks, we provide the agent sparse reward signals (a positive reward for reaching the goal only) instead of expert data, so they are RL rather than IL tasks. We use HPPO proposed in Section \ref{frame} and initialize it with the transferred options (i.e., HPPO-init). To be more convincing, we use two other SOTA HRL and RL algorithms -- \textbf{DAC} \cite{DBLP:conf/nips/ZhangW19a} and \textbf{PPO} \cite{DBLP:journals/corr/SchulmanWDRK17}, as baselines. In Figure \ref{fig:6(e)} and \ref{fig:6(f)}, we plot the episodic reward change in the training process of each algorithm, where the solid line and shadow represent the mean and standard deviation of the performance across the 4 different goals in each scenario. We can see that the reuse of options significantly accelerate the learning process and the newly proposed HRL algorithm performs much better than the baselines. Note that the other algorithms are trained for more episodes since they do not adopt the transferred options. We show that, in scenarios for which we do not have expert data or dense rewards, we can make use of the basic skills learned from expert demonstrations for similar task scenarios to effectively aid the learning, which provides a manner to bridge IL and RL.


\section{Conclusion and Discussion}

In this paper, we propose MH-AIRL to learn a hierarchical policy that can be adopted to perform a class of tasks, based on a mixture of multi-task unannotated expert data. We evaluate our algorithm on a series of challenging robotic multi-task settings. The results show that the multi-task hierarchical policies trained with MH-AIRL perform significantly better than the monotonic policies learned with SOTA Multi-task/Meta IL baselines. Further, with MH-AIRL, the agent can capture the subtask structures in each task and form a skill for each subtask. The basic skills can be reused for different tasks in that distribution to improve the expert data efficiency, and can even be transferred to more distinct tasks out of the distribution to solve long-timescale sparse-reward RL problems.

The primary limitation of our study is the inherent complexity of the overall framework, which comprises five networks as depicted in Figure \ref{fig:3}. This complexity arises from our algorithm's integration of AIRL, context-based Meta IL, and the option framework. This amalgamation introduces certain challenges in the training process, particularly in determining the optimal number of training iterations for each network within each learning episode. After careful fine-tuning, we established a training iteration ratio of 1:3:10 for the discriminator, hierarchical policy, and variational posteriors, respectively. Despite this complexity, our evaluations across a wide variety of tasks utilized a consistent set of hyperparameters, showing the robustness of our approach.


\bibliography{main}
\bibliographystyle{icml2023}

\newpage
\appendix
\onecolumn
\section{Appendix on the Background and Related Works}

\subsection{AIRL Framework to Solve Equation \ref{equ:6}} \label{c-airl}

For each task $C$, we need to recover the task-specific reward function $\mathcal{R}_{\vartheta}(S,A|C)$ and policy $\pi(A|S,C)$ based on the corresponding expert trajectories $\tau_{E} \sim \pi_{E}(\cdot|C)$ which can be solved by AIRL as mentioned in Section \ref{airl}. Thus, we have the following objective functions for training, which is a simple extension of AIRL \cite{DBLP:conf/nips/GhasemipourGZ19, DBLP:conf/nips/YuYFE19}:
\begin{equation} \label{equ:a12}
\begin{aligned}
        \mathop{\min}_{\vartheta}\mathbb{E}_{C} \left[-\mathbb{E}_{\tau_{E} \sim \pi_{E}(\cdot|C)}\left[\mathop{\sum}_{t=0}^{T-1}\log D_{\vartheta}(S_t,A_t|C)\right] - \mathbb{E}_{\tau \sim \pi(\cdot|C)}\left[\mathop{\sum}_{t=0}^{T-1}\log(1-D_{\vartheta}(S_t,A_t|C))\right]\right]
\end{aligned}
\end{equation}
\begin{equation} \label{equ:a13}
\begin{aligned}
        \mathop{\max}_{\pi}\mathbb{E}_{C} \left[\mathbb{E}_{\tau \sim \pi(\cdot|C)}\left[\mathop{\sum}_{t=0}^{T-1}\log D_{\vartheta}(S_t,A_t|C) - \log(1-D_{\vartheta}(S_t,A_t|C))\right]\right]
\end{aligned}
\end{equation}
where $D_{\vartheta}(S,A|C)=\exp(f_{\vartheta}(S,A|C))/[\exp(f_{\vartheta}(S,A|C))+\pi(A|S,C)]$.

\subsection{Implementation of the Hierarchical Policy in the One-step Option Model} \label{mha}

In this section, we give out the detailed structure design of the hierarchical policy introduced in Section \ref{option}, i.e., $\pi_{\theta}(Z|S, Z')$ and $\pi_{\phi}(A|S,Z)$, which is proposed in \cite{li2020skill}. This part is not our contribution, so we only provide the details for the purpose of implementation.

As mentioned in Section \ref{option}, the structure design is based on the  Multi-Head Attention (MHA) mechanism \cite{DBLP:conf/nips/VaswaniSPUJGKP17}. An attention function can be described as mapping a query, i.e., $q \in \mathbb{R}^{d_k}$, and a set of key-value pairs, i.e., $K=[k_1 \cdots k_n]^T \in \mathbb{R}^{n \times d_k}$ and $V=[v_1 \cdots v_n]^T \in \mathbb{R}^{n \times d_v}$, to an output. The output is computed as a weighted sum of the values, where the weight assigned to each value is computed by a compatibility function of the query with the corresponding key. To be specific:
\begin{equation} \label{equ:a8}  
\begin{aligned}
    Attention(q, K, V) = \mathop{\sum}_{i=1}^n\left[\frac{\exp(q \cdot k_i)}{\mathop{\sum}_{j=1}^n \exp(q \cdot k_j)} \times v_i \right]
\end{aligned}
\end{equation}
where $q,K,V$ are learnable parameters, $\frac{\exp(q \cdot k_i)}{\mathop{\sum}_{j=1}^n \exp(q \cdot k_j)}$ represents the attention weight that the model should pay to item $i$. In MHA, the query and key-value pairs are first linearly projected $h$ times to get $h$ different queries, keys and values. Then, an attention function is performed on each of these projected versions of queries, keys and values in parallel to get $h$ outputs which are then be concatenated and linearly projected to acquire the final output. The whole process can be represented as Equation \ref{equ:a9}, where $W_i^q \in \mathbb{R}^{d_k \times d_k}, W_i^K \in \mathbb{R}^{d_k \times d_k}, W_i^V \in \mathbb{R}^{d_v \times d_v}, W^O \in \mathbb{R}^{nd_v \times d_v}$ are the learnable parameters. 
\begin{equation} \label{equ:a9}  
\begin{aligned}
    MHA(q,K,V)=Concat(head_1, \cdots, head_h)W^O,\  head_i=Attention(qW_i^q, KW_i^K, VW_i^V)
\end{aligned}
\end{equation}

In this work, the option is represented as an $N$-dimensional one-hot vector, where $N$ denotes the total number of options to learn. The high-level policy $\pi_{\theta}(Z|S, Z')$ has the structure shown as:
\begin{equation} \label{equ:a10}  
\begin{aligned}
    q=linear(Concat[S, W_C^TZ']),\ dense_Z = MHA(q, W_C, W_C),\ Z \sim Categorical(\cdot|dense_Z)
\end{aligned}
\end{equation}
$W_C \in \mathbb{R}^{N \times E}$ is the option context matrix of which the $i$-th row represents the context embedding of the option $i$. $W_C$ is also used as the key and value matrix for the MHA, so $d_k=d_v=E$ in this case. Note that $W_C$ is only updated in the MHA module. Intuitively, $\pi_{\theta}(Z|S, Z')$ attends to all the option context embeddings in $W_C$ according to $S$ and $Z'$. If $Z'$ still fits $S$, $\pi_{\theta}(Z|S, Z')$ will assign a larger attention weight to $Z'$ and thus has a tendency to continue with it; otherwise, a new skill with better compatibility will be sampled.

As for the low-level policy $\pi_{\phi}(A|S,Z)$, it has the following structure:
\begin{equation} \label{equ:a11}  
\begin{aligned}
    dense_A = MLP(S, W_C^TZ),\ A \sim Categorical/Gaussian(\cdot|dense_A)
\end{aligned}
\end{equation}
where $MLP$ represents a multilayer perceptron, $A$ follows a categorical distribution for the discrete case or a gaussian distribution for the continuous case. The context embedding corresponding to $Z$, i.e., $W_C^TZ$, instead of $Z$ only, is used as input of $\pi_{\phi}$ since it can encode multiple properties of the option $Z$ \cite{DBLP:conf/nips/KosiorekSTH19}.

\section{Appendix on the Hierarchical Latent context Structure}

\subsection{A Lower Bound of the Directed Information Objective} \label{dilb}

In this section, we give out the derivation of a lower bound of the directed information from the trajectory sequence $X_{0:T}$ to the local latent context sequence $Z_{0:T}$ conditioned on the global latent context $C$, i.e., $I(X_{0:T} \rightarrow Z_{0:T}|C)$ as follows:
\begin{equation} \label{equ:a1}  
\begin{aligned}
    &I(X_{0:T} \rightarrow Z_{0:T}|C) =  \mathop{\sum}_{t=1}^{T}\left[I(X_{0:t};Z_{t}|Z_{0:t-1}, C)\right] \\
    &= \mathop{\sum}_{t=1}^{T}\left[H(Z_t|Z_{0:t-1}, C)-H(Z_t|X_{0:t},Z_{0:t-1}, C)\right] \\
    &\geq \mathop{\sum}_{t=1}^{T}\left[H(Z_t|X_{0:t-1}, Z_{0:t-1}, C)-H(Z_t|X_{0:t},Z_{0:t-1}, C)\right] \\
    &= \mathop{\sum}_{t=1}^{T} [H(Z_t|X_{0:t-1}, Z_{0:t-1}, C) + \\
    & \qquad\quad\mathop{\sum}_{\substack{X_{0:t}, C,\\ Z_{0:t-1}}}P(X_{0:t}, Z_{0:t-1}, C)\mathop{\sum}_{Z_t}P(Z_t|X_{0:t}, Z_{0:t-1}, C)\log P(Z_t|X_{0:t}, Z_{0:t-1}, C)]
\end{aligned}
\end{equation}
In Equation \ref{equ:a1}, $I(Var_1;Var_2|Var_3)$ denotes the conditional mutual information, $H(Var_1|Var_2)$ denotes the conditional entropy, and the inequality holds because of the basic property related to conditional entropy: increasing conditioning
cannot increase entropy \cite{galvin2014three}. $H(Z_t|X_{0:t-1}, Z_{0:t-1}, C)$ is the entropy of the high-level policy $\pi_{\theta}(Z_t|S_{t-1}, Z_{t-1})$, where the other variables in $X_{0:t-1}, Z_{0:t-1}$ are neglected due to the one-step Markov assumption, and more convenient to obtain. Further, the second term in the last step can be processed as follows:
\begin{equation} \label{equ:a2}  
\begin{aligned}
    &\mathop{\sum}_{Z_t}P(Z_t|X_{0:t}, Z_{0:t-1}, C)\log P(Z_t|X_{0:t}, Z_{0:t-1}, C)\\
    &= \mathop{\sum}_{Z_t}P(Z_t|X_{0:t}, Z_{0:t-1}, C)\left[\log\frac{P(Z_t|X_{0:t}, Z_{0:t-1}, C)}{P_{\omega}(Z_t|X_{0:t}, Z_{0:t-1}, C)} + \log P_{\omega}(Z_t|X_{0:t}, Z_{0:t-1}, C)\right]\\
    &= D_{KL}(P(\cdot|X_{0:t}, Z_{0:t-1}, C)||P_{\omega}(\cdot|X_{0:t}, Z_{0:t-1}, C)) + \mathop{\sum}_{Z_t}P(Z_t|X_{0:t}, Z_{0:t-1}, C)\log P_{\omega}(Z_t|X_{0:t}, Z_{0:t-1}, C)\\
    &\geq \mathop{\sum}_{Z_t}P(Z_t|X_{0:t}, Z_{0:t-1}, C)\log P_{\omega}(Z_t|X_{0:t}, Z_{0:t-1}, C)
\end{aligned}
\end{equation}
where $D_{KL}(\cdot)$ denotes the Kullback-Leibler (KL) Divergence which is non-negative \cite{cover1999elements}, $P_{\omega}(Z_t|X_{0:t}, Z_{0:t-1}, C)$ is a variational estimation of the posterior distribution of $Z_t$ given $X_{0:t}$ and $Z_{0:t-1}$, i.e., $P(Z_t|X_{0:t}, Z_{0:t-1}, C)$, which is modeled as a recurrent neural network with the parameter set $\omega$ in our work. Based on Equation \ref{equ:a1} and \ref{equ:a2}, we can obtain a lower bound of $I(X_{0:T} \rightarrow Z_{0:T}|C)$ denoted as $L^{DI}$:
\begin{equation} \label{equ:a3}  
\begin{aligned}
    L^{DI}=\mathop{\sum}_{t=1}^{T}[\mathop{\sum}_{\substack{X_{0:t}, C,\\ Z_{0:t}}}P(X_{0:t}, Z_{0:t}, C)\log P_{\omega}(Z_t|X_{0:t}, Z_{0:t-1}, C) + H(Z_t|X_{0:t-1}, Z_{0:t-1}, C)]
\end{aligned}
\end{equation}
Note that the joint distribution $P(X_{0:t}, Z_{0:t}, C)$ has a recursive definition as follows:
\begin{equation} \label{equ:a4}  
\begin{aligned}
    &P(X_{0:t}, Z_{0:t}, C) = prior(C)P(X_{0:t}, Z_{0:t}|C)\\
    &=prior(C)P(X_{t}|X_{0:t-1}, Z_{0:t}, C)P(Z_{t}|X_{0:t-1}, Z_{0:t-1}, C)P(X_{0:t-1}, Z_{0:t-1}|C)
\end{aligned}
\end{equation}
\begin{equation} \label{equ:a5}  
\begin{aligned}
    P(X_{0}, Z_{0}|C) = P((S_{0},A_{-1}), Z_{0}|C) = \mu(S_{0}|C)
\end{aligned}
\end{equation}
where $\mu(S_{0}|C)$ denotes the distribution of the initial states for task $C$. Equation \ref{equ:a5} holds because $A_{-1}$ and $Z_{0}$ are dummy variables which are only for simplifying notations and never executed and set to be constant across different tasks. Based on Equation \ref{equ:a4} and \ref{equ:a5}, we can get:
\begin{equation} \label{equ:a6}  
\begin{aligned}
    &P(X_{0:t}, Z_{0:t}, C) = prior(C)\mu(S_{0}|C)\mathop{\prod}_{i=1}^{t}P(Z_{i}|X_{0:i-1}, Z_{0:i-1}, C)P(X_{i}|X_{0:i-1}, Z_{0:i}, C)\\
    &=prior(C)\mu(S_{0}|C)\mathop{\prod}_{i=1}^{t}P(Z_{i}|X_{0:i-1}, Z_{0:i-1}, C)P((S_i, A_{i-1})|X_{0:i-1}, Z_{0:i}, C)\\
    &=prior(C)\mu(S_{0}|C)\mathop{\prod}_{i=1}^{t}P(Z_{i}|X_{0:i-1}, Z_{0:i-1}, C)P( A_{i-1}|X_{0:i-1}, Z_{0:i}, C)\mathcal{P}(S_i|S_{i-1},A_{i-1}, C) \\
    &=prior(C)\mu(S_{0}|C)\mathop{\prod}_{i=1}^{t}\pi_{\theta}(Z_{i}|S_{i-1}, Z_{i-1}, C)\pi_{\phi}( A_{i-1}|S_{i-1}, Z_{i}, C)\mathcal{P}(S_i|S_{i-1},A_{i-1}, C)
\end{aligned}
\end{equation}
In Equation \ref{equ:a6}, $prior(C)$ is the known prior distribution of the task context $C$, $\mathcal{P}(S_i|S_{i-1},A_{i-1}, C)$ is the transition dynamic of task $C$, $P(Z_{i}|X_{0:i-1}, Z_{0:i-1}, C)$ and $P( A_{i-1}|X_{0:i-1}, Z_{0:i}, C)$ can be replaced with $\pi_{\theta}$ and $\pi_{\phi}$, respectively, due to the one-step Markov assumption. 


To sum up, we can adopt the high-level policy, low-level policy and variational posterior to get an estimation of the lower bound of the directed information objective through Monte Carlo sampling \cite{sutton2018reinforcement} according to Equation \ref{equ:a3} and \ref{equ:a6}, which can then be used to optimize the three networks.

\subsection{A Lower Bound of the Mutual Information Objective} \label{milb}

In this section, we give out the derivation of a lower bound of the mutual information between the trajectory sequence $X_{0:T}$ and its corresponding task context $C$, i.e., $I(X_{0:T};C)$.
\begin{equation} \label{equ:a7}  
\begin{aligned}
    &I(X_{0:T};C) = H(C) - H(C|X_{0:T})\\
    &=H(C)+\mathop{\sum}_{X_{0:T}}P(X_{0:T})\mathop{\sum}_{C}P(C|X_{0:T})\log P(C|X_{0:T})\\
    &=H(C)+\mathop{\sum}_{X_{0:T}}P(X_{0:T})\mathop{\sum}_{C}P(C|X_{0:T})\log\frac{P(C|X_{0:T})}{P_{\psi}(C|X_{0:T})}+\mathop{\sum}_{X_{0:T}, C}P(X_{0:T}, C)\log P_{\psi}(C|X_{0:T})\\
    &=H(C)+\mathop{\sum}_{X_{0:T}}P(X_{0:T})D_{KL}(P(\cdot|X_{0:T}||P_{\psi}(\cdot|X_{0:T}))+\mathop{\sum}_{X_{0:T}, C}P(X_{0:T}, C)\log P_{\psi}(C|X_{0:T})\\
    &\geq H(C)+\mathop{\sum}_{X_{0:T}, C}P(X_{0:T}, C)\log P_{\psi}(C|X_{0:T})\\
    &=H(C)+\mathop{\sum}_{C}prior(C)\mathop{\sum}_{X_{0:T}}P(X_{0:T}|C)\log P_{\psi}(C|X_{0:T})\\
    &=H(C)+\mathop{\sum}_{C}prior(C)\mathop{\sum}_{X_{0:T}, Z_{0:T}}P(X_{0:T}, Z_{0:T}|C)\log P_{\psi}(C|X_{0:T})
\end{aligned}
\end{equation}
In Equation \ref{equ:a7}, $H(\cdot)$ denotes the entropy, $prior(C)$ denotes the known prior distribution of the task context $C$, $P(X_{0:T}, Z_{0:T}|C)$ can be calculated with Equation \ref{equ:a6} by setting $t=T$, and $P_{\psi}(C|X_{0:T})$ is a variational estimation of the posterior distribution $P(C|X_{0:T})$ which is implemented as a  recurrent neural network with the parameter set $\psi$. Note that the inequality holds because the KL-Divergence, i.e., $D_{KL}(\cdot)$, is non-negative.

\subsection{The Analogy with the VAE Framework} \label{rnn-imp}

Variational Autoencoder (VAE) \cite{DBLP:journals/corr/KingmaW13} learns a probabilistic \textbf{encoder} $P_{\eta}(V|U)$ and \textbf{decoder} $P_{\xi}(U|V)$ which map between data $U$ and latent variables $V$ by optimizing the evidence lower bound (ELBO) on the marginal distribution $P_{\xi}(U)$, assuming the prior distributions $P^U(\cdot)$ and $P^V(\cdot)$ over the data and latent variables respectively. The authors of \cite{DBLP:conf/iclr/HigginsMPBGBML17} extend the VAE approach by including a parameter $\beta$ to control the capacity of the latent $V$, of which the ELBO is:
\begin{equation} \label{equ:100}
\begin{aligned}
    \mathop{\max}_{\eta, \xi}\mathop{\mathbb{E}}_{\substack{U\sim P^U(\cdot) \\ V\sim P_{\eta}(\cdot|U)}}\left[\log P_{\xi}(U|V)-\beta D_{KL}(P_{\eta}(V|U)||P^V(V))\right]
\end{aligned}
\end{equation}
The first term can be viewed as the reconstruction accuracy of the data $U$ from $V$, and the second term works as a regularizer for the distribution of the latent variables $V$, where $D_{KL}$ denotes the KL Divergence \cite{cover1999elements}. VAE can efficiently solve the posterior inference problem for datasets with continuous latent variables where the true posterior is intractable, through fitting an approximate inference model $P_{\xi}$ (i.e., the variational posterior). The variational lower bound, i.e., ELBO, can be straightforwardly optimized using standard stochastic gradient methods, e.g., SGD \cite{bottou2010large}.

\begin{figure*}[t]
\centering
\includegraphics[width=5.5in, height=1.0in]{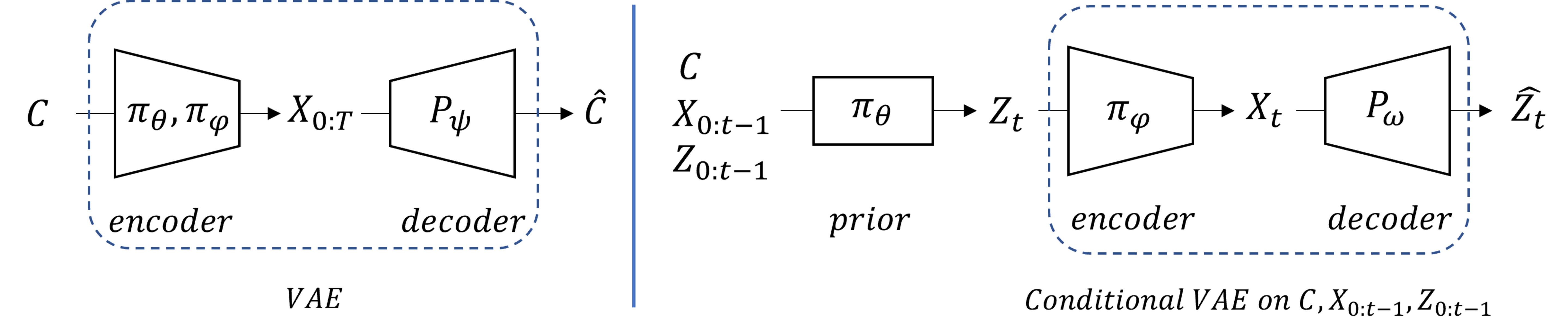}
\caption{The analogy of our learning framework with the VAE structure.}
\label{fig:2}
\end{figure*}

As shown in Figure \ref{fig:2}, the optimization of $L^{MI}$ (Equation \ref{equ:14}) can be viewed as using $\pi_{\theta}$ and $\pi_{\phi}$ as the encoder and $P_{\psi}$ as the decoder and then minimizing the reconstruction error of $C$ from $X_{0:T}$, and the regularizer term in Equation \ref{equ:100} is neglected (i.e., $\beta = 0$). As for the optimization of $L^{DI}$ (Equation \ref{equ:14}), at each timestep $t$, $\pi_{\phi}$ and $P_{\omega}$ form a conditional VAE between $Z_t$ and $X_t$, which is conditioned on the history information and task code, i.e., $(X_{0:t-1},Z_{0:t-1},C)$, with the prior distribution of $Z_t$ provided by $\pi_{\theta}$. Compared with the VAE objective (i.e., Equation \ref{equ:100}), $\pi_{\phi}$ and $P_{\omega}$ in $L^{DI}$ work as the encoder and decoder respectively; $\pi_{\theta}$ provides the prior, which corresponds to $P^U(\cdot)$.


Both $P_{\psi}$ and $P_{\omega}$ use sequential data as input and thus are implemented with RNN. The variational posterior for the task code, i.e., $P_{\psi}(C|X_{0:T})$ takes the trajectory $X_{0:T}$ as input and is implemented as a bidirectional GRU \cite{DBLP:journals/corr/abs-1908-04332} to make sure that both the beginning and end of the trajectory are equally important. On the other hand, the variational posterior for the local latent code, i.e., $P_{\omega}(Z_t|X_{0:t}, Z_{0:t-1}, C)$, is modeled as $P_{\omega}(Z_t|X_t, Z_{t-1}, C, h_{t-1})$, where $h_{t-1}$ is the internal hidden state of an RNN. $h_{t-1}$ is recursively maintained with the time series using the GRU rule, i.e., $h_{t-1}=GRU(X_{t-1},Z_{t-2},h_{t-2})$, to embed the history information in the trajectory, i.e., $X_{0:t-1}$ and $Z_{0:t-2}$. Note that the RNN-based posterior has been used and justified in the process for sequential data \cite{DBLP:conf/nips/ChungKDGCB15}.

\newpage

\section{Appendix on Hierarchical AIRL}

\subsection{Derivation of the MLE Objective} \label{mle-obj}
In Equation \ref{equ:a31}, $Z_0$ is a dummy variable which is assigned before the episode begins and never executed. It's implemented as a constant across different episodes, so we have $P(S_0,Z_0|C)=P(S_0|C)=\mu(S_0|C)$, where $\mu(\cdot|C)$ denotes the initial state distribution for task $C$. On the other hand, we have $P(S_{t+1},Z_{t+1}|S_t, Z_t, Z_{t+1}, A_{t}, C)=P(Z_{t+1}|S_t, Z_t, Z_{t+1}, A_{t}, C)P(S_{t+1}|S_t, Z_t, Z_{t+1}, A_{t}, C)=\mathcal{P}(S_{t+1}|S_t, A_{t}, C)$, since the transition dynamic $\mathcal{P}$ is irrelevant to the local latent codes $Z$ and only related the task context $C$.
\begin{equation} \label{equ:a31}
\begin{aligned}
&P_{\vartheta}(X_{0:T}, Z_{0:T}|C)\propto \mu(\widetilde{S}_0|C)\mathop{\prod}_{t=0}^{T-1}\mathcal{P}(\widetilde{S}_{t+1}|\widetilde{S}_{t}, \widetilde{A}_{t}, C)\exp(\mathcal{R}_{\vartheta}(\widetilde{S}_t, \widetilde{A}_t|C)) \\
&=P(S_0,Z_0|C)\mathop{\prod}_{t=0}^{T-1}P(S_{t+1},Z_{t+1}|S_t, Z_t, Z_{t+1}, A_{t}, C)\exp(\mathcal{R}_{\vartheta}(S_t, Z_{t}, Z_{t+1}, A_{t}|C))\\
&=\mu(S_0|C)\mathop{\prod}_{t=0}^{T-1}\mathcal{P}(S_{t+1}|S_t, A_{t}, C)\exp(\mathcal{R}_{\vartheta}(S_t, Z_{t}, Z_{t+1}, A_{t}|C))\\
\end{aligned}
\end{equation}

\subsection{Justification of the Objective Function Design in Equation \ref{equ:10}} \label{just}

In this section, we prove that by optimizing the objective functions shown in Equation \ref{equ:10}, we can get the solution of the MLE problem shown as Equation \ref{equ:8}, i.e., the task-conditioned hierarchical reward function and policy of the expert.

In Appendix A of \cite{DBLP:journals/corr/abs-1710-11248}, they show that the discriminator objective (the first equation in \ref{equ:10}) is equivalent to the MLE objective (Equation \ref{equ:8}) where $f_{\vartheta}$ serves as $R_{\vartheta}$, when $D_{KL}(\pi(\tau)||\pi_E(\tau))$ is minimized. The same conclusion can be acquired by simply replacing $\{S_t,A_t,\tau\}$ with $\{(S_t, Z_t),(Z_{t+1}, A_t),(X_{0:T}, Z_{0:T})\}$, i.e., the extended definition of the state, action and trajectory, in the original proof, which we don't repeat here. Then, we only need to prove that $\mathbb{E}_{C}\left[D_{KL}(\pi_{\theta, \phi}(X_{0:T}, Z_{0:T}|C)||\pi_E(X_{0:T}, Z_{0:T}|C))\right]$ can be minimized through the second equation in \ref{equ:10}:
\begin{equation} \label{equ:a32}
\begin{aligned}
    &\mathop{\max}_{\theta, \phi}\mathbb{E}_{C \sim prior(\cdot), (X_{0:T}, Z_{0:T}) \sim \pi_{\theta, \phi}(\cdot|C)}\mathop{\sum}_{t=0}^{T-1}R_{IL}^{t}\\
    &=\mathop{\max}_{\theta, \phi}\mathop{\mathbb{E}}_{C, X_{0:T}, Z_{0:T}}\left[\mathop{\sum}_{t=0}^{T-1}\log D_{\vartheta}(S_t,Z_t,Z_{t+1},A_t|C) - \log(1-D_{\vartheta}(S_t,Z_t,Z_{t+1},A_t|C))\right]\\
    &=\mathop{\max}_{\theta, \phi}\mathop{\mathbb{E}}_{C, X_{0:T}, Z_{0:T}}\left[\mathop{\sum}_{t=0}^{T-1}f_{\vartheta}(S_t,Z_t,Z_{t+1},A_t|C) - \log\pi_{\theta, \phi}(Z_{t+1},A_t|S_t,Z_t,C)\right]\\
    &=\mathop{\max}_{\theta, \phi}\mathop{\mathbb{E}}_{C, X_{0:T}, Z_{0:T}}\left[\mathop{\sum}_{t=0}^{T-1}f_{\vartheta}(S_t,Z_t,Z_{t+1},A_t|C) - \log(\pi_{\theta}(Z_{t+1}|S_t,Z_t,C)\pi_{\phi}(A_t|S_t,Z_{t+1}, C))\right]\\
    &=\mathop{\max}_{\theta, \phi}\mathop{\mathbb{E}}_{C, X_{0:T}, Z_{0:T}}\left[\log\frac{\prod_{t=0}^{T-1}\exp(f_{\vartheta}(S_t,Z_t,Z_{t+1},A_t|C))}{\prod_{t=0}^{T-1}\pi_{\theta}(Z_{t+1}|S_t,Z_t,C)\pi_{\phi}(A_t|S_t,Z_{t+1},C)}\right]\\
    &\iff\mathop{\max}_{\theta, \phi}\mathop{\mathbb{E}}_{C, X_{0:T}, Z_{0:T}}\left[\log\frac{\prod_{t=0}^{T-1}\exp(f_{\vartheta}(S_t,Z_t,Z_{t+1},A_t|C))/Z_{\vartheta}^C}{\prod_{t=0}^{T-1}\pi_{\theta}(Z_{t+1}|S_t,Z_t,C)\pi_{\phi}(A_t|S_t,Z_{t+1},C)}\right]\\
\end{aligned}
\end{equation}
Note that $Z_{\vartheta}^C=\sum_{X_{0:T}, Z_{0:T}}\widehat{P}_{\vartheta}(X_{0:T}, Z_{0:T}|C)$ (defined in Equation \ref{equ:8}) is the normalized function parameterized with $\vartheta$, so the introduction of $Z_{\vartheta}^C$ will not influence the optimization with respect to $\theta$ and $\phi$ and the equivalence at the last step holds. Also, the second equality shows that the task-conditioned hierarchical policy is recovered by optimizing an entropy-regularized policy objective where $f_{\vartheta}$ serves as $R_{\vartheta}$. Further, we have:
\begin{equation} \label{equ:a33}
\begin{aligned}
    &\mathop{\max}_{\theta, \phi}\mathop{\mathbb{E}}_{C, X_{0:T}, Z_{0:T}}\left[\log\frac{\prod_{t=0}^{T-1}\exp(f_{\vartheta}(S_t,Z_t,Z_{t+1},A_t|C))/Z_{\vartheta}^C}{\prod_{t=0}^{T-1}\pi_{\theta}(Z_{t+1}|S_t,Z_t,C)\pi_{\phi}(A_t|S_t,Z_{t+1},C)}\right]\\
    &=\mathop{\max}_{\theta, \phi}\mathop{\mathbb{E}}_{C, X_{0:T}, Z_{0:T}}\left[\log\frac{\mu(S_0|C)\prod_{t=0}^{T-1}\mathcal{P}(S_{t+1}|S_t, A_{t}, C)\prod_{t=0}^{T-1}\exp(f_{\vartheta}(S_t,Z_t,Z_{t+1},A_t|C))/Z_{\vartheta}^C}{\mu(S_0|C)\prod_{t=0}^{T-1}\mathcal{P}(S_{t+1}|S_t, A_{t}, C)\prod_{t=0}^{T-1}\pi_{\theta}(Z_{t+1}|S_t,Z_t,C)\pi_{\phi}(A_t|S_t,Z_{t+1},C)}\right]\\
    &=\mathop{\max}_{\theta, \phi}\mathbb{E}_{C \sim prior(\cdot), (X_{0:T}, Z_{0:T}) \sim \pi_{\theta, \phi}(\cdot|C)}\left[\log\frac{\pi_{E}(X_{0:T},Z_{0:T}|C)}{\pi_{\theta, \phi}(X_{0:T},Z_{0:T}|C)}\right]\\
    &=\mathop{\max}_{\theta, \phi}\mathbb{E}_{C \sim prior(\cdot)}\left[-D_{KL}(\pi_{\theta, \phi}(X_{0:T}, Z_{0:T}|C)||\pi_E(X_{0:T}, Z_{0:T}|C))\right]\\
    & \iff\mathop{\min}_{\theta, \phi}\mathbb{E}_{C \sim prior(\cdot)}\left[D_{KL}(\pi_{\theta, \phi}(X_{0:T}, Z_{0:T}|C)||\pi_E(X_{0:T}, Z_{0:T}|C))\right]
\end{aligned}
\end{equation}
where the second equality holds because of the definition of $\pi_{E}$ (Equation \ref{equ:8} with $f_{\vartheta}$ serving as $R_{\vartheta}$) and $\pi_{\theta, \phi}$ (Equation \ref{equ:a15}).

\subsection{Justification of the EM-style Adaption} \label{em-airl}

Given only a dataset of expert trajectories, i.e., $D_{E} \triangleq \{X_{0:T}\}$, we can still maximize the likelihood estimation $\mathbb{E}_{X_{0:T} \sim D_{E}}\left[\log P_{\vartheta}(X_{0:T})\right]$ through an EM-style adaption: (We use $X_{0:T}, C, Z_{0:T}$ instead of $X_{0:T}^E, C_E, Z_{0:T}^E$ for simplicity.)
\begin{equation} \label{equ:a34}
\begin{aligned}
    &\mathbb{E}_{X_{0:T} \sim D_{E}}\left[\log P_{\vartheta}(X_{0:T})\right]=\mathbb{E}_{X_{0:T} \sim D_{E}}\left[\log\left[\mathop{\sum}_{C, Z_{0:T}}P_{\vartheta}(X_{0:T},C,Z_{0:T})\right]\right]\\
    &=\mathbb{E}_{X_{0:T} \sim D_{E}}\left[\log\left[\mathop{\sum}_{C, Z_{0:T}}\frac{P_{\vartheta}(X_{0:T},C,Z_{0:T})}{P_{\overline{\vartheta}}(C,Z_{0:T}|X_{0:T})}P_{\overline{\vartheta}}(C,Z_{0:T}|X_{0:T})\right]\right]\\
    &=\mathbb{E}_{X_{0:T} \sim D_{E}}\left[\log\left[\mathbb{E}_{(C, Z_{0:T})\sim P_{\overline{\vartheta}}(\cdot|X_{0:T})}\frac{P_{\vartheta}(X_{0:T},C,Z_{0:T})}{P_{\overline{\vartheta}}(C,Z_{0:T}|X_{0:T})}\right]\right]\\
    &\geq \mathbb{E}_{X_{0:T} \sim D_{E}}\left[\mathbb{E}_{(C, Z_{0:T})\sim P_{\overline{\vartheta}}(\cdot|X_{0:T})}\log\frac{P_{\vartheta}(X_{0:T},C,Z_{0:T})}{P_{\overline{\vartheta}}(C,Z_{0:T}|X_{0:T})}\right]\\
    &= \mathbb{E}_{X_{0:T} \sim D_{E}, C \sim P_{\overline{\psi}}(\cdot|X_{0:T}), Z_{0:T} \sim P_{\overline{\omega}}(\cdot|X_{0:T},C)}\left[\log\frac{P_{\vartheta}(X_{0:T},C,Z_{0:T})}{P_{\overline{\vartheta}}(C,Z_{0:T}|X_{0:T})}\right]\\
    &= \mathbb{E}_{X_{0:T}, C , Z_{0:T}}\left[\log P_{\vartheta}(X_{0:T},C,Z_{0:T})\right] - \mathbb{E}_{X_{0:T}, C , Z_{0:T}}\left[\log P_{\overline{\vartheta}}(C,Z_{0:T}|X_{0:T})\right]\\
    &= \mathbb{E}_{X_{0:T}, C , Z_{0:T}}\left[\log P_{\vartheta}(X_{0:T},Z_{0:T}|C)\right] - \mathbb{E}_{X_{0:T}, C , Z_{0:T}}\left[-\log prior(C) + \log P_{\overline{\vartheta}}(C,Z_{0:T}|X_{0:T})\right]\\
\end{aligned}
\end{equation}
where we adopt the Jensen's inequality \cite{jensen1906fonctions} in the 4-th step. Also, we note that $P_{\overline{\psi}, \overline{\omega}}(C, Z_{0:T}|X_{0:T})$ provides a posterior distribution of $(C, Z_{0:T})$, which corresponds to the generating process led by the hierarchical policy. As justified in \ref{just}, the hierarchical policy is trained with the reward function parameterized with $\overline{\vartheta}$. Thus, the hierarchical policy is a function of $\overline{\vartheta}$, and the network $P_{\overline{\psi}, \overline{\omega}}$ corresponding to the hierarchical policy provides a posterior distribution related to the parameter set $\overline{\vartheta}$, i.e., $(C, Z_{0:T})\sim P_{\overline{\vartheta}}(\cdot|X_{0:T}) \iff C \sim P_{\overline{\psi}}(\cdot|X_{0:T}), Z_{0:T} \sim P_{\overline{\omega}}(\cdot|X_{0:T},C)$, due to which the 5-th step holds. Note that $\overline{\vartheta}, \overline{\psi}, \overline{\omega}$ denote the parameters $\vartheta, \psi, \omega$ before being updated in the M step.

In the second equality of Equation \ref{equ:a34}, we introduce the sampled global and local latent codes in the E step as discussed in Section \ref{h-airl}. Then, in the M step, we optimize the objectives shown in Equation \ref{equ:14} and \ref{equ:10} for iterations, by replacing the samples in the first term of Equation \ref{equ:10} with $(X_{0:T}, C, Z_{0:T})$ collected in the E step.  This is equivalent to solve the MLE problem: $\mathop{\max}_{\vartheta}\mathbb{E}_{X_{0:T} \sim D_{E}, C \sim P_{\overline{\psi}}(\cdot|X_{0:T}), Z_{0:T}\sim P_{\overline{\omega}}(\cdot|X_{0:T},C)}\left[\log P_{\vartheta}(X_{0:T},Z_{0:T}|C)\right]$, which is to maximize a lower bound of the original objective, i.e., $\mathbb{E}_{X_{0:T} \sim D_{E}}\left[\log P_{\vartheta}(X_{0:T})\right]$, as shown in the last step of Equation \ref{equ:a34}. Thus, the original objective can be optimized through this EM procedure. Note that the second term in the last step is a function of the old parameter $\overline{\vartheta}$ so that it can be overlooked when optimizing with respect to $\vartheta$.

\subsection{State-only Adaption of H-AIRL} \label{so-airl}
In AIRL \cite{DBLP:journals/corr/abs-1710-11248}, they propose a two-component design for the discriminator as follows:
\begin{equation} \label{equ:b1}
\begin{aligned}
    f_{\vartheta, \zeta}(S_t, S_{t+1}) = g_{\vartheta}(S_t) + \gamma h_{\zeta}(S_{t+1}) - h_{\zeta}(S_t)
\end{aligned}
\end{equation}
where $\gamma$ is the discount factor in MDP. Based on $f_{\vartheta, \zeta}(S_t, S_{t+1})$, they can further get $D_{\vartheta, \zeta}(S_t, S_{t+1})$ which is used in Equation \ref{equ:3} for AIRL training. As proved in \cite{DBLP:journals/corr/abs-1710-11248}, $g_{\vartheta}$, $h_{\zeta}$ and $f_{\vartheta, \zeta}$ can recover the true reward, value and advantage function, respectively, under deterministic environments with a state-only ground truth reward. With this state-only design, the recovered reward function is disentangled from the dynamics of the environment in which it was trained, so that it can be directly transferred to environments with different transition dynamics, i.e., $\mathcal{P}$, for the policy training. Moreover, the additional shaping term $h_{\zeta}$ helps mitigate the effects of unwanted shaping on the reward approximator $g_{\vartheta}$ \cite{DBLP:conf/icml/NgHR99}. This design can also be adopted to H-AIRL (Equation \ref{equ:10}) by redefining Equation \ref{equ:b1} on the extended state space (first defined in Section \ref{h-airl}):
\begin{equation} \label{equ:b2}
\begin{aligned}
    f_{\vartheta, \zeta}(\widetilde{S}_t, \widetilde{S}_{t+1}|C) &= g_{\vartheta}(\widetilde{S}_t|C) + \gamma h_{\zeta}(\widetilde{S}_{t+1}|C) - h_{\zeta}(\widetilde{S}_t|C)\\
    & = g_{\vartheta}(S_t, Z_t|C) + \gamma h_{\zeta}(S_{t+1}, Z_{t+1}|C) - h_{\zeta}(S_t, Z_t|C)
\end{aligned}
\end{equation}
In this way, we can recover a hierarchical reward function conditioned on the task context $C$, i.e., $g_{\vartheta}(S_t, Z_t|C)$, which avoids unwanted shaping and is robust enough to be directly applied in a new task with different dynamic transition distribution from $prior(C)$. The proof can be done by simply replacing the state $S$ in the original proof (Appendix C of \cite{DBLP:journals/corr/abs-1710-11248}) with its extended definition $\widetilde{S}$, so we don't repeat it here.

\section{The Proposed Actor-Critic Algorithm for Training}


\subsection{Gradients of the Mutual Information Objective Term} \label{mi-gra}

The objective function related to the mutual information:
\begin{equation} \label{equ:a14}  
\begin{aligned}
    L^{MI}=\mathop{\sum}_{C}prior(C)\mathop{\sum}_{X_{0:T}, Z_{0:T}}P(X_{0:T}, Z_{0:T}|C)\log P_{\psi}(C|X_{0:T})
\end{aligned}
\end{equation}
After introducing the one-step Markov assumption to Equation \ref{equ:a6}, we can calculate $P(X_{0:T}, Z_{0:T}|C)$ as Equation \ref{equ:a15}, where $\pi_{\theta}$ and $\pi_{\phi}$ represent the hierarchical policy in the one-step option framework.
\begin{equation} \label{equ:a15}  
\begin{aligned}
    P(X_{0:T}, Z_{0:T}|C)=\mu(S_{0}|C)\mathop{\prod}_{t=1}^{T}\pi_{\theta}(Z_{t}|S_{t-1}, Z_{t-1}, C)\pi_{\phi}( A_{t-1}|S_{t-1}, Z_{t}, C)\mathcal{P}(S_t|S_{t-1},A_{t-1}, C)
\end{aligned}
\end{equation}
First, the gradient with respect to $\psi$ is straightforward as Equation \ref{equ:a35}, which can be optimized as a standard likelihood maximization problem.
\begin{equation} \label{equ:a35}  
\begin{aligned}
    \nabla_{\psi}L^{MI}=\mathop{\sum}_{C}prior(C)\mathop{\sum}_{X_{0:T}, Z_{0:T}}P(X_{0:T}, Z_{0:T}|C)\nabla_{\psi}\log P_{\psi}(C|X_{0:T})
\end{aligned}
\end{equation}
Now we give out the derivation of $\nabla_{\theta}L^{MI}$:
\begin{equation} \label{equ:a16}  
\begin{aligned}
    &\nabla_{\theta}L^{MI}=\mathop{\sum}_{C}prior(C)\mathop{\sum}_{X_{0:T}, Z_{0:T}}\nabla_{\theta}P_{\theta,\phi}(X_{0:T}, Z_{0:T}|C)\log P_{\psi}(C|X_{0:T})\\
    &=\mathop{\sum}_{C}prior(C)\mathop{\sum}_{X_{0:T}, Z_{0:T}}P_{\theta,\phi}(X_{0:T}, Z_{0:T}|C)\nabla_{\theta}\log P_{\theta,\phi}(X_{0:T}, Z_{0:T}|C)\log P_{\psi}(C|X_{0:T})\\
    &=\mathop{\mathbb{E}}_{\substack{C,X_{0:T},\\Z_{0:T}}}\left[\nabla_{\theta}\log P_{\theta,\phi}(X_{0:T}, Z_{0:T}|C)\log P_{\psi}(C|X_{0:T})\right]\\
    &=\mathop{\mathbb{E}}_{\substack{C,X_{0:T},\\Z_{0:T}}}\left[\mathop{\sum}_{t=1}^{T}\nabla_{\theta}\log\pi_{\theta}(Z_{t}|S_{t-1}, Z_{t-1}, C)\log P_{\psi}(C|X_{0:T})\right]
\end{aligned}
\end{equation}
where the last equality holds because of Equation \ref{equ:a15}. With similar derivation as above, we have:
\begin{equation} \label{equ:a17}  
\begin{aligned}
    \nabla_{\phi}L^{MI}=\mathop{\mathbb{E}}_{\substack{C,X_{0:T},\\Z_{0:T}}}\left[\mathop{\sum}_{t=1}^{T}\nabla_{\phi}\log\pi_{\phi}( A_{t-1}|S_{t-1}, Z_{t}, C)\log P_{\psi}(C|X_{0:T})\right]
\end{aligned}
\end{equation}

\subsection{Gradients of the Directed Information Objective Term} \label{di-gra}

Next, we give out the derivation of the gradients related to the directed information objective term, i.e., $L^{DI}$. We denote the two terms in Equation \ref{equ:a3} as $L^{DI}_{1}$ and $L^{DI}_{2}$ respectively. Then, we have $\nabla_{\theta,\phi}L^{DI}=\nabla_{\theta,\phi}L^{DI}_{1}+\nabla_{\theta,\phi}L^{DI}_{2}$. The derivations are as follows:
\begin{equation} \label{equ:a18}
    \begin{aligned} 
    &\nabla_{\theta}L^{DI}_{1}=\mathop{\sum}_{t=1}^{T} \mathop{\sum}_{C}prior(C)\mathop{\sum}_{X_{0:t}, Z_{0:t}}\nabla_{\theta}P_{\theta,\phi}(X_{0:t}, Z_{0:t}|C)\log P_{\omega}(Z_t|X_{0:t}, Z_{0:t-1}, C)\\
    &=\mathop{\sum}_{t=1}^{T} \mathop{\sum}_{C}prior(C)\mathop{\sum}_{X_{0:t}, Z_{0:t}}P_{\theta,\phi}(X_{0:t}, Z_{0:t}|C)\mathop{\sum}_{i=1}^{t}\nabla_{\theta}\log\pi_{\theta}(Z_{i}|S_{i-1}, Z_{i-1}, C)\log P_{\omega}^{t}\\
    &=\mathop{\sum}_{t=1}^{T} \mathop{\sum}_{C}prior(C)\mathop{\sum}_{X_{0:t}, Z_{0:t}}\mathop{\sum}_{\substack{X_{t+1:T},\\ Z_{t+1:T}}}P_{\theta,\phi}(X_{0:T}, Z_{0:T}|C)\mathop{\sum}_{i=1}^{t}\nabla_{\theta}\log\pi_{\theta}(Z_{i}|S_{i-1}, Z_{i-1}, C)\log P_{\omega}^{t}\\
    &=\mathop{\sum}_{t=1}^{T} \mathop{\sum}_{C}prior(C)\mathop{\sum}_{X_{0:T}, Z_{0:T}}P_{\theta,\phi}(X_{0:T}, Z_{0:T}|C)\mathop{\sum}_{i=1}^{t}\nabla_{\theta}\log\pi_{\theta}(Z_{i}|S_{i-1}, Z_{i-1}, C)\log P_{\omega}^{t}\\
    &=\mathop{\sum}_{C}prior(C)\mathop{\sum}_{X_{0:T}, Z_{0:T}}P_{\theta,\phi}(X_{0:T}, Z_{0:T}|C)\mathop{\sum}_{t=1}^{T}\log P_{\omega}^t \mathop{\sum}_{i=1}^{t}\nabla_{\theta}\log\pi_{\theta}(Z_{i}|S_{i-1}, Z_{i-1}, C)\\
    &=\mathop{\sum}_{C}prior(C)\mathop{\sum}_{X_{0:T}, Z_{0:T}}P_{\theta,\phi}(X_{0:T}, Z_{0:T}|C)\mathop{\sum}_{i=1}^{T}\nabla_{\theta}\log\pi_{\theta}(Z_{i}|S_{i-1}, Z_{i-1}, C) \mathop{\sum}_{t=i}^{T}\log P_{\omega}^t\\
    &=\mathop{\mathbb{E}}_{\substack{C,X_{0:T},\\Z_{0:T}}}\left[\mathop{\sum}_{i=1}^{T}\nabla_{\theta}\log\pi_{\theta}(Z_{i}|S_{i-1}, Z_{i-1}, C) \mathop{\sum}_{t=i}^{T}\log P_{\omega}(Z_t|X_{0:t}, Z_{0:t-1}, C)\right]\\
    &=\mathop{\mathbb{E}}_{\substack{C,X_{0:T},\\Z_{0:T}}}\left[\mathop{\sum}_{t=1}^{T}\nabla_{\theta}\log\pi_{\theta}(Z_{t}|S_{t-1}, Z_{t-1}, C) \mathop{\sum}_{i=t}^{T}\log P_{\omega}(Z_i|X_{0:i}, Z_{0:i-1}, C)\right]
\end{aligned}
\end{equation}
where $P_{\omega}^t=P_{\omega}(Z_t|X_{0:t}, Z_{0:t-1}, C)$ for simplicity. The second equality in Equation \ref{equ:a18} holds following the same derivation in Equation \ref{equ:a16}. 
Then, the gradient related to $L^{DI}_{2}$ is:
\begin{equation} \label{equ:a37}
    \begin{aligned} 
    &\nabla_{\theta}L^{DI}_{2}=\nabla_{\theta}\mathop{\sum}_{t=1}^{T}H(Z_t|X_{0:t-1},Z_{0:t-1},C)\\
    &=-\nabla_{\theta}[\mathop{\sum}_{t=1}^{T} \mathop{\sum}_{C}prior(C)\mathop{\sum}_{X_{0:t-1}, Z_{0:t}}P_{\theta,\phi}(X_{0:t-1}, Z_{0:t}|C)\log P(Z_t|X_{0:t-1},Z_{0:t-1},C)]\\
    &=-\nabla_{\theta}[\mathop{\sum}_{t=1}^{T} \mathop{\sum}_{C}prior(C)\mathop{\sum}_{X_{0:t-1}, Z_{0:t}}P_{\theta,\phi}(X_{0:t-1}, Z_{0:t}|C)\log\pi_{\theta}(Z_t|S_{t-1},Z_{t-1},C)]\\
    &=-\nabla_{\theta}[\mathop{\sum}_{C}prior(C)\mathop{\sum}_{X_{0:T}, Z_{0:T}}P_{\theta,\phi}(X_{0:T}, Z_{0:T}|C)\mathop{\sum}_{t=1}^{T}\log\pi_{\theta}(Z_t|S_{t-1},Z_{t-1},C)]\\
    &=-[\mathop{\sum}_{C}prior(C)\mathop{\sum}_{X_{0:T}, Z_{0:T}}\nabla_{\theta}P_{\theta,\phi}(X_{0:T}, Z_{0:T}|C)\mathop{\sum}_{t=1}^{T}\log\pi_{\theta}(Z_t|S_{t-1},Z_{t-1},C)+\\
    &\qquad\ \mathop{\sum}_{C}prior(C)\mathop{\sum}_{X_{0:T}, Z_{0:T}}P_{\theta,\phi}(X_{0:T}, Z_{0:T}|C)\mathop{\sum}_{t=1}^{T}\nabla_{\theta}\log\pi_{\theta}(Z_t|S_{t-1},Z_{t-1},C)]
\end{aligned}
\end{equation}
\begin{equation} \label{equ:a37a}
    \begin{aligned} 
    &=-\mathop{\mathbb{E}}_{\substack{C,X_{0:T},\\Z_{0:T}}}\left[\mathop{\sum}_{t=1}^{T}\nabla_{\theta}\log\pi_{\theta}(Z_{t}|S_{t-1}, Z_{t-1}, C)\left[\mathop{\sum}_{i=1}^{T}\log\pi_{\theta}(Z_i|S_{i-1},Z_{i-1},C)+1\right]\right]\\
    &=-\mathop{\mathbb{E}}_{\substack{C,X_{0:T},\\Z_{0:T}}}\left[\mathop{\sum}_{t=1}^{T}\nabla_{\theta}\log\pi_{\theta}(Z_{t}|S_{t-1}, Z_{t-1}, C)\mathop{\sum}_{i=t}^{T}\log\pi_{\theta}(Z_i|S_{i-1},Z_{i-1},C)\right]\\
\end{aligned}
\end{equation}

The third equality holds because we adopt the one-step Markov assumption, i.e., the conditional probability distribution of a random variable depends only on its parent nodes in the probabilistic graphical model (shown as Figure \ref{fig:1}). The fourth equality holds out of similar derivation as steps 2-4 in Equation \ref{equ:a18}. The last equality can be obtained with Equation \ref{equ:a22} in the next section, where we prove that any term which is from $\mathop{\sum}_{i=1}^{T}\log\pi_{\theta}(Z_i|S_{i-1},Z_{i-1},C)+1$ and not a function of $Z_{t}$ will not influence the gradient calculation in Equation \ref{equ:a37} and \ref{equ:a37a}.

With similar derivations, we have:
\begin{equation} \label{equ:a19}  
\begin{aligned}
    &\nabla_{\phi}L^{DI}_{1}=\mathop{\mathbb{E}}_{\substack{C,X_{0:T},\\Z_{0:T}}}\left[\mathop{\sum}_{t=1}^{T}\nabla_{\phi}\log\pi_{\phi}( A_{t-1}|S_{t-1}, Z_{t}, C) \mathop{\sum}_{i=t}^{T}\log P_{\omega}(Z_i|X_{0:i}, Z_{0:i-1}, C)\right]
\end{aligned}
\end{equation}
\begin{equation} \label{equ:a38}
    \begin{aligned} 
    \nabla_{\phi}L^{DI}_{2}=-\mathop{\mathbb{E}}_{\substack{C,X_{0:T},\\Z_{0:T}}}\left[\mathop{\sum}_{t=1}^{T}\nabla_{\phi}\log\pi_{\phi}( A_{t-1}|S_{t-1}, Z_{t}, C)\mathop{\sum}_{i=t}^{T}\log\pi_{\theta}(Z_i|S_{i-1},Z_{i-1},C)\right]\\
\end{aligned}
\end{equation}

As for the gradient with respect to $\omega$, it can be computed with:
\begin{equation} \label{equ:a36}
    \begin{aligned} 
    \nabla_{\omega}L^{DI}=\nabla_{\omega}L^{DI}_{1}=\mathop{\sum}_{t=1}^{T} \mathop{\sum}_{C}prior(C)\mathop{\sum}_{X_{0:t}, Z_{0:t}}P_{\theta,\phi}(X_{0:t}, Z_{0:t}|C)\nabla_{\omega}\log P_{\omega}(Z_t|X_{0:t}, Z_{0:t-1}, C)
\end{aligned}
\end{equation}
Still, for each timestep $t$, it's a standard likelihood maximization problem and can be optimized through SGD.

\subsection{Gradients of the Imitation Learning Objective Term} \label{il-gra}

We consider the imitation learning objective term $L^{IL}$, i.e., the trajectory return shown as:
\begin{equation} \label{equ:a20}  
\begin{aligned}
    L^{IL}=\mathop{\sum}_{C}prior(C)\mathop{\sum}_{X_{0:T}, Z_{0:T}}P_{\theta, \phi}(X_{0:T}, Z_{0:T}|C)\mathop{\sum}_{i=0}^{T-1}R_{IL}(S_i,Z_i,Z_{i+1},A_i|C)
\end{aligned}
\end{equation}
Following the similar derivation with Equation \ref{equ:a16}, we can get:
\begin{equation} \label{equ:a21}  
\begin{aligned}
    \nabla_{\theta}L^{IL}=\mathop{\mathbb{E}}_{\substack{C,X_{0:T},\\Z_{0:T}}}\left[\mathop{\sum}_{t=1}^{T}\nabla_{\theta}\log\pi_{\theta}(Z_{t}|S_{t-1}, Z_{t-1}, C)\mathop{\sum}_{i=0}^{T-1}R_{IL}(S_i,Z_i,Z_{i+1},A_i|C)\right]
\end{aligned}
\end{equation}
Further, we note that for each $t\in\{1,\cdots,T\}$, $\forall i < t-1$, we have:
\begin{equation} \label{equ:a22}  
\begin{aligned}
    &\mathop{\mathbb{E}}_{\substack{C,X_{0:T},\\Z_{0:T}}}\left[\nabla_{\theta}\log\pi_{\theta}(Z_{t}|S_{t-1}, Z_{t-1}, C)R_{IL}(S_i,Z_i,Z_{i+1},A_i|C)\right]\\
    &=\mathop{\sum}_{C}prior(C)\mathop{\sum}_{X_{0:T}, Z_{0:T}}P_{\theta, \phi}(X_{0:T}, Z_{0:T}|C)\nabla_{\theta}\log\pi_{\theta}(Z_{t}|S_{t-1}, Z_{t-1}, C)R_{IL}(S_i,Z_i,Z_{i+1},A_i|C)\\
    &=\mathop{\sum}_{C}prior(C)\mathop{\sum}_{\substack{X_{0:t-1},\\ Z_{0:t}}}\mathop{\sum}_{\substack{X_{t:T},\\ Z_{t+1:T}}}P_{\theta, \phi}(X_{0:T}, Z_{0:T}|C)\nabla_{\theta}\log\pi_{\theta}(Z_{t}|S_{t-1}, Z_{t-1}, C)R_{IL}^i\\
    &=\mathop{\sum}_{C}prior(C)\mathop{\sum}_{\substack{X_{0:t-1},\\ Z_{0:t}}}P_{\theta, \phi}(X_{0:t-1}, Z_{0:t}|C)\nabla_{\theta}\log\pi_{\theta}(Z_{t}|S_{t-1}, Z_{t-1}, C)R_{IL}^i\\
\end{aligned}
\end{equation}
\begin{equation} \label{equ:a39}  
\begin{aligned}
&=\mathop{\sum}_{C}prior(C)\mathop{\sum}_{\substack{X_{0:t-1},\\ Z_{0:t-1}}}P_{\theta, \phi}(X_{0:t-1}, Z_{0:t-1}|C)\mathop{\sum}_{Z_t}\pi_{\theta}(Z_{t}|S_{t-1}, Z_{t-1}, C)\nabla_{\theta}\log\pi_{\theta}(Z_{t}|S_{t-1}, Z_{t-1}, C)R_{IL}^i\\
    &=\mathop{\sum}_{C}prior(C)\mathop{\sum}_{\substack{X_{0:t-1},\\ Z_{0:t-1}}}P_{\theta, \phi}(X_{0:t-1}, Z_{0:t-1}|C)R_{IL}^i\mathop{\sum}_{Z_t}\pi_{\theta}(Z_{t}|S_{t-1}, Z_{t-1}, C)\nabla_{\theta}\log\pi_{\theta}(Z_{t}|S_{t-1}, Z_{t-1}, C)\\
    &=\mathop{\sum}_{C}prior(C)\mathop{\sum}_{\substack{X_{0:t-1},\\ Z_{0:t-1}}}P_{\theta, \phi}(X_{0:t-1}, Z_{0:t-1}|C)R_{IL}^i\mathop{\sum}_{Z_t}\nabla_{\theta}\pi_{\theta}(Z_{t}|S_{t-1}, Z_{t-1}, C)\\
    &=\mathop{\sum}_{C}prior(C)\mathop{\sum}_{\substack{X_{0:t-1},\\ Z_{0:t-1}}}P_{\theta, \phi}(X_{0:t-1}, Z_{0:t-1}|C)R_{IL}^i\nabla_{\theta}\mathop{\sum}_{Z_t}\pi_{\theta}(Z_{t}|S_{t-1}, Z_{t-1}, C)\\
    &=\mathop{\sum}_{C}prior(C)\mathop{\sum}_{\substack{X_{0:t-1},\\ Z_{0:t-1}}}P_{\theta, \phi}(X_{0:t-1}, Z_{0:t-1}|C)R_{IL}(S_i,Z_i,Z_{i+1},A_i|C)\nabla_{\theta}1=0\\
\end{aligned}
\end{equation}
where $R_{IL}^i=R_{IL}(S_i,Z_i,Z_{i+1},A_i|C)$ for simplicity. We use the law of total probability in the third equality, which we also use in the later derivations. The fifth equality holds because $i<t-1$ and $R_{IL}(S_i,Z_i,Z_{i+1},A_i|C)$ is irrelevant to $Z_t$. Based on Equation \ref{equ:a21} and \ref{equ:a22}, we have:
\begin{equation} \label{equ:a23}  
\begin{aligned}
    \nabla_{\theta}L^{IL}=\mathop{\mathbb{E}}_{\substack{C,X_{0:T},\\Z_{0:T}}}\left[\mathop{\sum}_{t=1}^{T}\nabla_{\theta}\log\pi_{\theta}(Z_{t}|S_{t-1}, Z_{t-1}, C)\mathop{\sum}_{i=t-1}^{T-1}R_{IL}(S_i,Z_i,Z_{i+1},A_i|C)\right]
\end{aligned}
\end{equation}
With similar derivations, we can obtain:
\begin{equation} \label{equ:a24}  
\begin{aligned}
    \nabla_{\phi}L^{IL}=\mathop{\mathbb{E}}_{\substack{C,X_{0:T},\\Z_{0:T}}}\left[\mathop{\sum}_{t=1}^{T}\nabla_{\phi}\log\pi_{\phi}( A_{t-1}|S_{t-1}, Z_{t}, C)\mathop{\sum}_{i=t-1}^{T-1}R_{IL}(S_i,Z_i,Z_{i+1},A_i|C)\right]
\end{aligned}
\end{equation}

\subsection{The Overall Unbiased Gradient Estimator} \label{overall}

To sum up, the gradients with respect to $\theta$ and $\phi$ can be computed with $\nabla_{\theta, \phi}L=\nabla_{\theta, \phi}(\alpha_1 L^{MI} + \alpha_2 L^{DI} + \alpha_3 L^{IL})$, where $\alpha_{1:3}>0$ are the weights for each objective term and fine-tuned as hyperparameters. Combining Equation (\ref{equ:a16}, \ref{equ:a18}, \ref{equ:a37}, \ref{equ:a23}) and Equation (\ref{equ:a17}, \ref{equ:a19}, \ref{equ:a38}, \ref{equ:a24}), we have the actor-critic learning framework shown as Equation \ref{equ:a25}, except for the baseline terms, $b^{high}$ and $b^{low}$.

 Further, we claim that Equation \ref{equ:a25} provides unbiased estimation of the gradients with respect to $\theta$ and $\phi$. We proof this by showing that $\mathop{\mathbb{E}}\left[\mathop{\sum}_{t=1}^{T}\nabla_{\theta}\log\pi_{\theta}^tb^{high}(S_{t-1}, Z_{t-1}|C)\right]=\mathop{\mathbb{E}}\left[\mathop{\sum}_{t=1}^{T}\nabla_{\phi}\log\pi_{\phi}^tb^{low}(S_{t-1}, Z_{t}| C)\right]=0$, as follows:
 
\begin{align*}
    &\mathop{\mathbb{E}}_{\substack{C,X_{0:T},\\Z_{0:T}}}\left[\mathop{\sum}_{t=1}^{T}\nabla_{\theta}\log\pi_{\theta}(Z_{t}|S_{t-1}, Z_{t-1}, C)b^{high}(S_{t-1}, Z_{t-1}|C)\right]\\
    &=\mathop{\sum}_{C}prior(C)\mathop{\sum}_{X_{0:T}, Z_{0:T}}P_{\theta,\phi}(X_{0:T}, Z_{0:T}|C)\mathop{\sum}_{t=1}^{T}\nabla_{\theta}\log\pi_{\theta}(Z_{t}|S_{t-1}, Z_{t-1}, C)b^{high}(S_{t-1}, Z_{t-1}|C)\\
    &=\mathop{\sum}_{C}prior(C)\mathop{\sum}_{t=1}^{T}\mathop{\sum}_{X_{0:T}, Z_{0:T}}P_{\theta,\phi}(X_{0:T}, Z_{0:T}|C)\nabla_{\theta}\log\pi_{\theta}(Z_{t}|S_{t-1}, Z_{t-1}, C)b^{high}(S_{t-1}, Z_{t-1}|C)\\
    &=\mathop{\sum}_{C}prior(C)\mathop{\sum}_{t=1}^{T}\mathop{\sum}_{\substack{X_{0:t-1},Z_{0:t}}}P_{\theta,\phi}(X_{0:t-1}, Z_{0:t}|C)\nabla_{\theta}\log\pi_{\theta}(Z_{t}|S_{t-1}, Z_{t-1}, C)b^{high}(S_{t-1}, Z_{t-1}|C)\\
    \\
    &\mathop{\sum}_{\substack{X_{0:t-1},Z_{0:t}}}P_{\theta,\phi}(X_{0:t-1}, Z_{0:t}|C)\nabla_{\theta}\log\pi_{\theta}(Z_{t}|S_{t-1}, Z_{t-1}, C)b^{high}(S_{t-1}, Z_{t-1}|C)\\
    &=\mathop{\sum}_{\substack{X_{0:t-1},\\ Z_{0:t-1}}}P_{\theta,\phi}(X_{0:t-1}, Z_{0:t-1}|C)\mathop{\sum}_{Z_{t}}\pi_{\theta}(Z_{t}|S_{t-1}, Z_{t-1}, C)\nabla_{\theta}\log\pi_{\theta}(Z_{t}|S_{t-1}, Z_{t-1}, C)b^{high}(S_{t-1}, Z_{t-1}|C)\\
    &=\mathop{\sum}_{\substack{X_{0:t-1}, Z_{0:t-1}}}P_{\theta,\phi}(X_{0:t-1}, Z_{0:t-1}|C)b^{high}(S_{t-1}, Z_{t-1}|C)\mathop{\sum}_{Z_{t}}\nabla_{\theta}\pi_{\theta}(Z_{t}|S_{t-1}, Z_{t-1}, C)\\
    &=\mathop{\sum}_{\substack{X_{0:t-1}, Z_{0:t-1}}}P_{\theta,\phi}(X_{0:t-1}, Z_{0:t-1}|C)b^{high}(S_{t-1}, Z_{t-1}|C)\nabla_{\theta}1=0
\end{align*}
\begin{align*}
    &\mathop{\mathbb{E}}_{\substack{C,X_{0:T},\\Z_{0:T}}}\left[\mathop{\sum}_{t=1}^{T}\nabla_{\phi}\log\pi_{\phi}( A_{t-1}|S_{t-1}, Z_{t}, C)b^{low}(S_{t-1}, Z_{t}| C)\right]\\
    &=\mathop{\sum}_{C}prior(C)\mathop{\sum}_{X_{0:T}, Z_{0:T}}P_{\theta,\phi}(X_{0:T}, Z_{0:T}|C)\mathop{\sum}_{t=1}^{T}\nabla_{\phi}\log\pi_{\phi}( A_{t-1}|S_{t-1}, Z_{t}, C)b^{low}(S_{t-1}, Z_{t}| C)\\
    &=\mathop{\sum}_{C}prior(C)\mathop{\sum}_{t=1}^{T}\mathop{\sum}_{X_{0:T}, Z_{0:T}}P_{\theta,\phi}(X_{0:T}, Z_{0:T}|C)\nabla_{\phi}\log\pi_{\phi}( A_{t-1}|S_{t-1}, Z_{t}, C)b^{low}(S_{t-1}, Z_{t}| C)\\
    &=\mathop{\sum}_{C}prior(C)\mathop{\sum}_{t=1}^{T}\mathop{\sum}_{X_{0:t}, Z_{0:t}}P_{\theta,\phi}(X_{0:t}, Z_{0:t}|C)\nabla_{\phi}\log\pi_{\phi}( A_{t-1}|S_{t-1}, Z_{t}, C)b^{low}(S_{t-1}, Z_{t}| C)\\
    &=\mathop{\sum}_{C}prior(C)\mathop{\sum}_{t=1}^{T}\mathop{\sum}_{X_{0:t-1}, Z_{0:t}}P_{\theta,\phi}(X_{0:t-1}, Z_{0:t}|C)\mathop{\sum}_{X_t}P_{\phi}(X_t|X_{0:t-1}, Z_{0:t}, C)\cdot\\
    &\qquad \qquad \qquad \qquad \qquad \qquad \quad \ \ \nabla_{\phi}\log\pi_{\phi}( A_{t-1}|S_{t-1}, Z_{t}, C)b^{low}(S_{t-1}, Z_{t}| C)\\
    &=\mathop{\sum}_{C}prior(C)\mathop{\sum}_{t=1}^{T}\mathop{\sum}_{\substack{X_{0:t-1}, Z_{0:t}}}P_{\theta,\phi}(X_{0:t-1}, Z_{0:t}|C)\mathop{\sum}_{A_{t-1}}\pi_{\phi}(A_{t-1}|S_{t-1}, Z_{t}, C)\cdot\\
    &\qquad \qquad \qquad \qquad \qquad \qquad \quad \ \ \nabla_{\phi}\log\pi_{\phi}( A_{t-1}|S_{t-1}, Z_{t}, C)b^{low}(S_{t-1}, Z_{t}| C)\mathop{\sum}_{S_t}\mathcal{P}(S_t|S_{t-1}, A_{t-1},C)\\
    &=\mathop{\sum}_{C}prior(C)\mathop{\sum}_{t=1}^{T}\mathop{\sum}_{\substack{X_{0:t-1}, Z_{0:t}}}P_{\theta,\phi}(X_{0:t-1}, Z_{0:t}|C)b^{low}(S_{t-1}, Z_{t}| C)\mathop{\sum}_{A_{t-1}}\pi_{\phi}(A_{t-1}|S_{t-1}, Z_{t}, C)\cdot\\
    &\qquad \qquad \qquad \qquad \qquad \qquad \quad \ \ \nabla_{\phi}\log\pi_{\phi}( A_{t-1}|S_{t-1}, Z_{t}, C)\\
    &=\mathop{\sum}_{C}prior(C)\mathop{\sum}_{t=1}^{T}\mathop{\sum}_{\substack{X_{0:t-1}, Z_{0:t}}}P_{\theta,\phi}(X_{0:t-1}, Z_{0:t}|C)b^{low}(S_{t-1}, Z_{t}| C)\mathop{\sum}_{A_{t-1}}\nabla_{\phi}\pi_{\phi}( A_{t-1}|S_{t-1}, Z_{t}, C)\\
    &\qquad \qquad \qquad \qquad \qquad \qquad \quad \ \ \\
    &=\mathop{\sum}_{C}prior(C)\mathop{\sum}_{t=1}^{T}\mathop{\sum}_{\substack{X_{0:t-1}, Z_{0:t}}}P_{\theta,\phi}(X_{0:t-1}, Z_{0:t}|C)b^{low}(S_{t-1}, Z_{t}| C)\nabla_{\phi}\mathop{\sum}_{A_{t-1}}\pi_{\phi}( A_{t-1}|S_{t-1}, Z_{t}, C)\\
    &=\mathop{\sum}_{C}prior(C)\mathop{\sum}_{t=1}^{T}\mathop{\sum}_{\substack{X_{0:t-1},\\ Z_{0:t}}}P_{\theta,\phi}(X_{0:t-1}, Z_{0:t}|C)b^{low}(S_{t-1}, Z_{t}| C)\nabla_{\phi}1=0\\
\end{align*}

\newpage

\subsection{Illustrations of Interactions among Networks in MH-AIRL} \label{illu}

\begin{algorithm*}[t]
	\caption{Multi-task Hierarchical Adversarial Inverse Reinforcement Learning (MH-AIRL)}\label{alg:1}
	\begin{algorithmic}[1]
	    \STATE \textbf{Input:} Prior distribution of the task variable $prior(C)$, expert demonstrations $\{X_{0:T}^E\}$ (If the task or option annotations, i.e., $\{C_E\}$ or $\{Z_{0:T}^E\}$, are provided, the corresponding estimation in Step 6 is not required.)
	    \STATE \textbf{Initialize} the hierarchical policy $\pi_{\theta}$ and $\pi_{\phi}$, discriminator $f_{\vartheta}$, posteriors for the task context $P_{\psi}$ and option choice $P_{\omega}$
		\FOR {$each\ training\ episode$}
		    \STATE \textbf{Generate} $M$ trajectories $\{(C, X_{0:T}, Z_{0:T})\}$ by sampling the task $C\sim prior(\cdot)$ and then exploring it with $\pi_{\theta}$ and $\pi_{\phi}$
		    \STATE \textbf{Update} $P_{\psi}$ and $P_{\omega}$ by minimizing $L^{MI}$ and $L^{DI}$ (Eq. \ref{equ:17}) using SGD with $\{(C, X_{0:T}, Z_{0:T})\}$
		    \STATE \textbf{Estimate} the expert global and local latent codes with $P_{\psi}$ and $P_{\omega}$, i.e., $C_{E} \sim P_{\psi}(\cdot|X_{0:T}^E)$, $Z_{0:T}^E \sim P_{\omega}(\cdot|X_{0:T}^E, C_{E})$
			\STATE \textbf{Update} $f_{\vartheta}$ by minimizing the cross entropy loss in Eq. \ref{equ:10} based on $\{(C, X_{0:T}, Z_{0:T})\}$ and $\{(C_E, X_{0:T}^E, Z_{0:T}^E)\}$
			\STATE \textbf{Train} $\pi_{\theta}$ and $\pi_{\phi}$ by HPPO, i.e., Eq. \ref{equ:a25}, based on $\{(C, X_{0:T}, Z_{0:T})\}$ and $f_{\vartheta}$ which defines $D_{\vartheta}$ and $R_{IL}$
		\ENDFOR
	\end{algorithmic} 
\end{algorithm*}

\begin{wrapfigure}{l}{6.1cm}
\centering
\includegraphics[width=2.2in, height=1.55in]{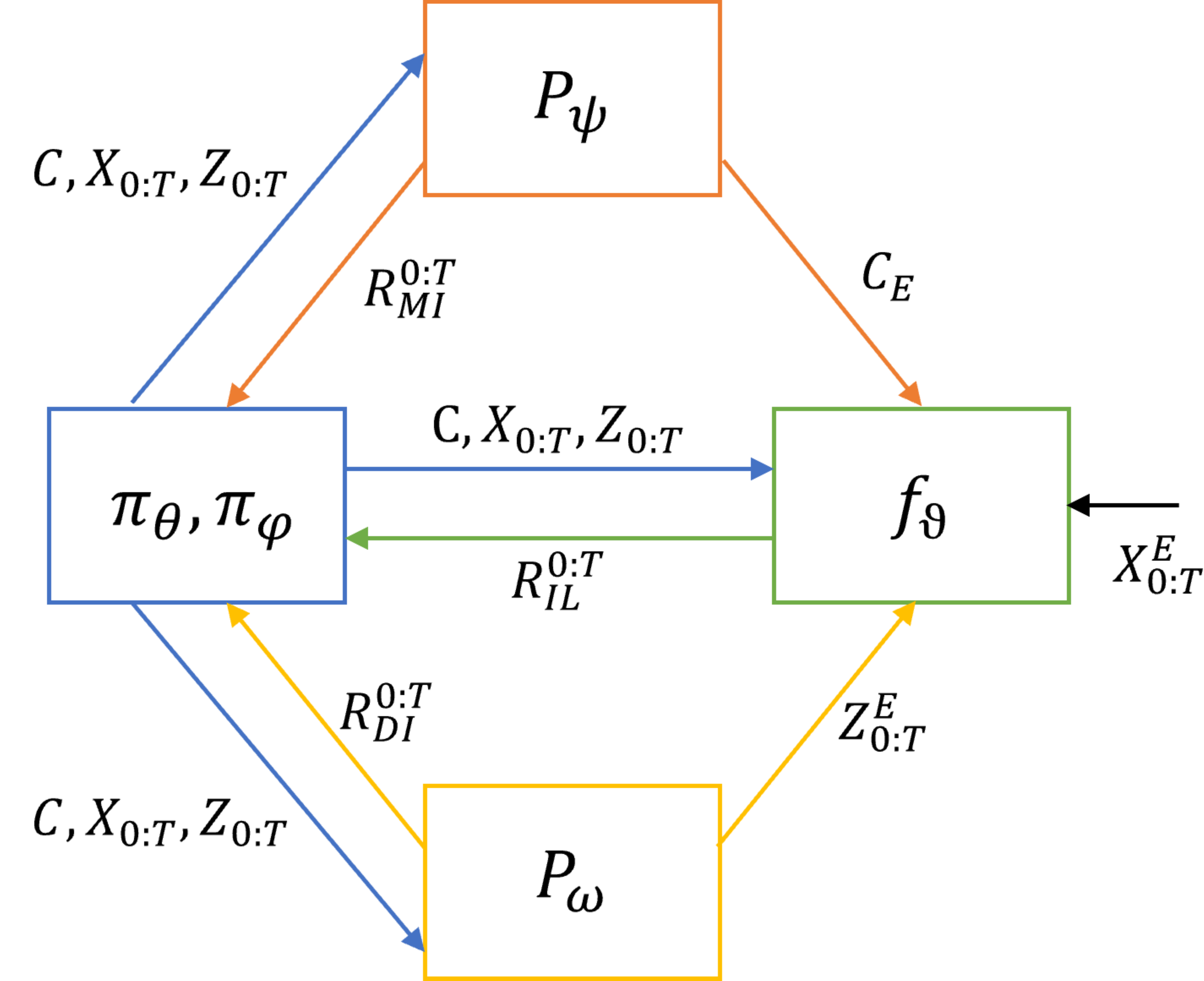}
\caption{Interactions among the five networks in our learning system.}
\label{fig:3} 
\end{wrapfigure}

There are in total five networks to learn in our system: the high-level policy $\pi_{\theta}$, low-level policy $\pi_{\phi}$, discriminator $f_{\vartheta}$, variational posteriors for the task context $P_{\psi}$ and option context $P_{\omega}$. Algorithm \ref{alg:1} shows in details how to coordinate their training process. To be more intuitive, we provide Figure \ref{fig:3} for illustrating the interactions among them. $P_{\psi}$ and $P_{\omega}$ are trained with the trajectories (i.e., $\{(C, X_{0:T}, Z_{0:T})\}$) generated by the hierarchical policy $\pi_{\theta, \phi}$, and can provide the reward signals $R_{MI}^{0:T}$ and $R_{DI}^{0:T}$ for training $\pi_{\theta, \phi}$, which are defined as $\alpha_1 \log P_{\psi}(C|X_{0:T})$ and $\alpha_2 \log\frac{P_{\omega}(Z_i|X^{i}, Z^{i-1}, C)}{ \pi_{\theta}(Z_i|S_{i-1},Z_{i-1},C)}$ ($i\in\{1, \cdots, T\}$) in Equation \ref{equ:a25}, respectively.
On the other hand, the discriminator $f_{\vartheta}$ is trained to distinguish the expert demonstrations $\{(C_E, X_{0:T}^E, Z_{0:T}^E)\}$ and generated samples $\{(C, X_{0:T}, Z_{0:T})\}$, where $C_E$ and $\{Z_{0:T}^E\}$ can be estimated from $P_{\psi}$ and $P_{\omega}$ if not provided. Then, the AIRL reward term $R_{IL}^{0:T}$ can be obtained based on the output of $f_{\vartheta}$. Last, the hierarchical policy $\pi_{\theta, \phi}$ can be trained by maximizing the return defined with $R_{MI}^{0:T}$, $R_{DI}^{0:T}$, and $R_{IL}^{0:T}$ (i.e., Eq. \ref{equ:a25}).

\newpage

\section{Appendix on Evaluation Results} 

\subsection{Plots of the Ablation Study} \label{ab-plots}
\begin{figure*}[htbp]
\centering
\subfigure[HalfCheetah-MultiVel]{
\label{fig:5(a)} 
\includegraphics[width=2.4in, height=1.2in]{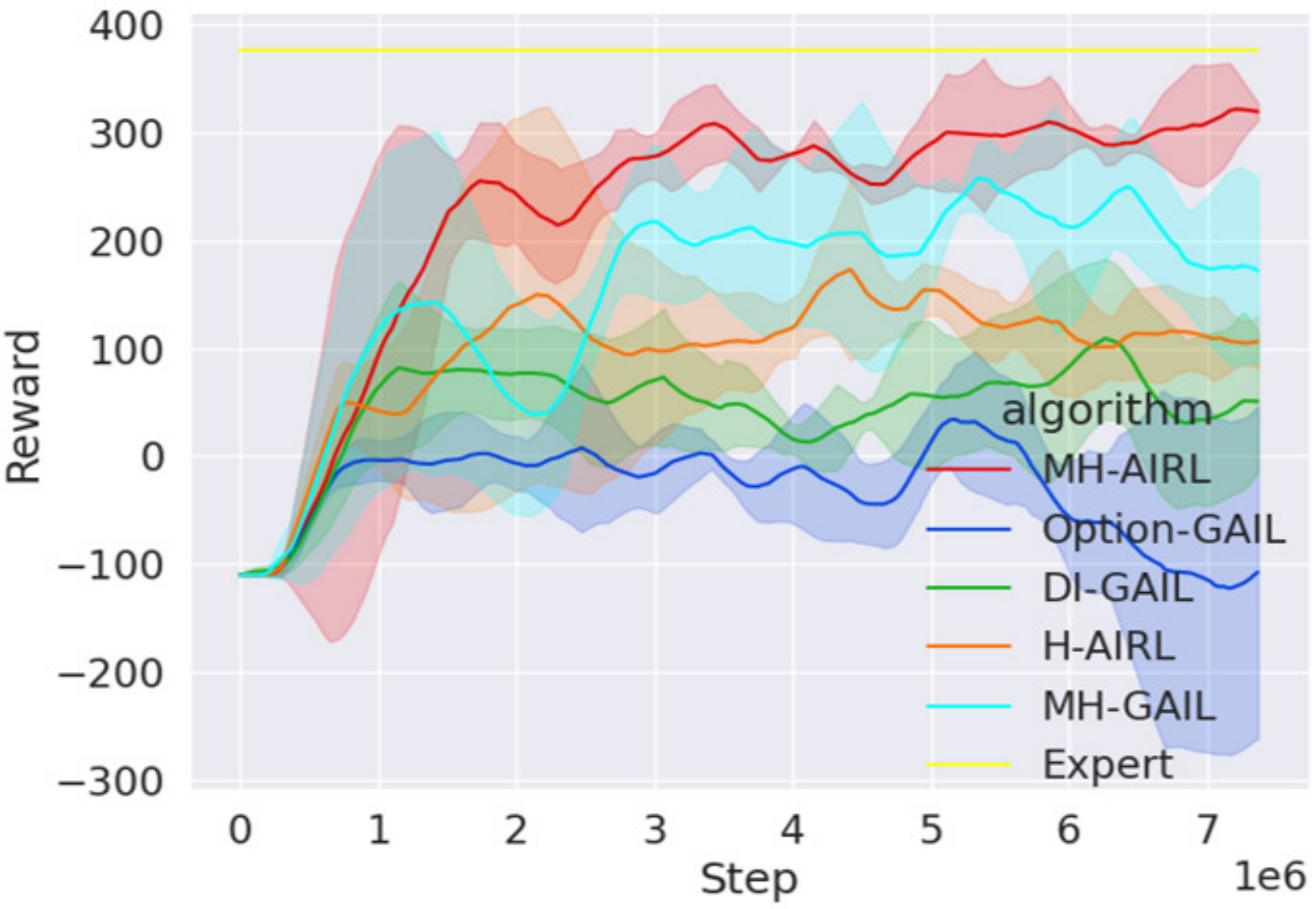}}
\subfigure[Walker-RandParam]{
\label{fig:5(b)} 
\includegraphics[width=2.4in, height=1.2in]{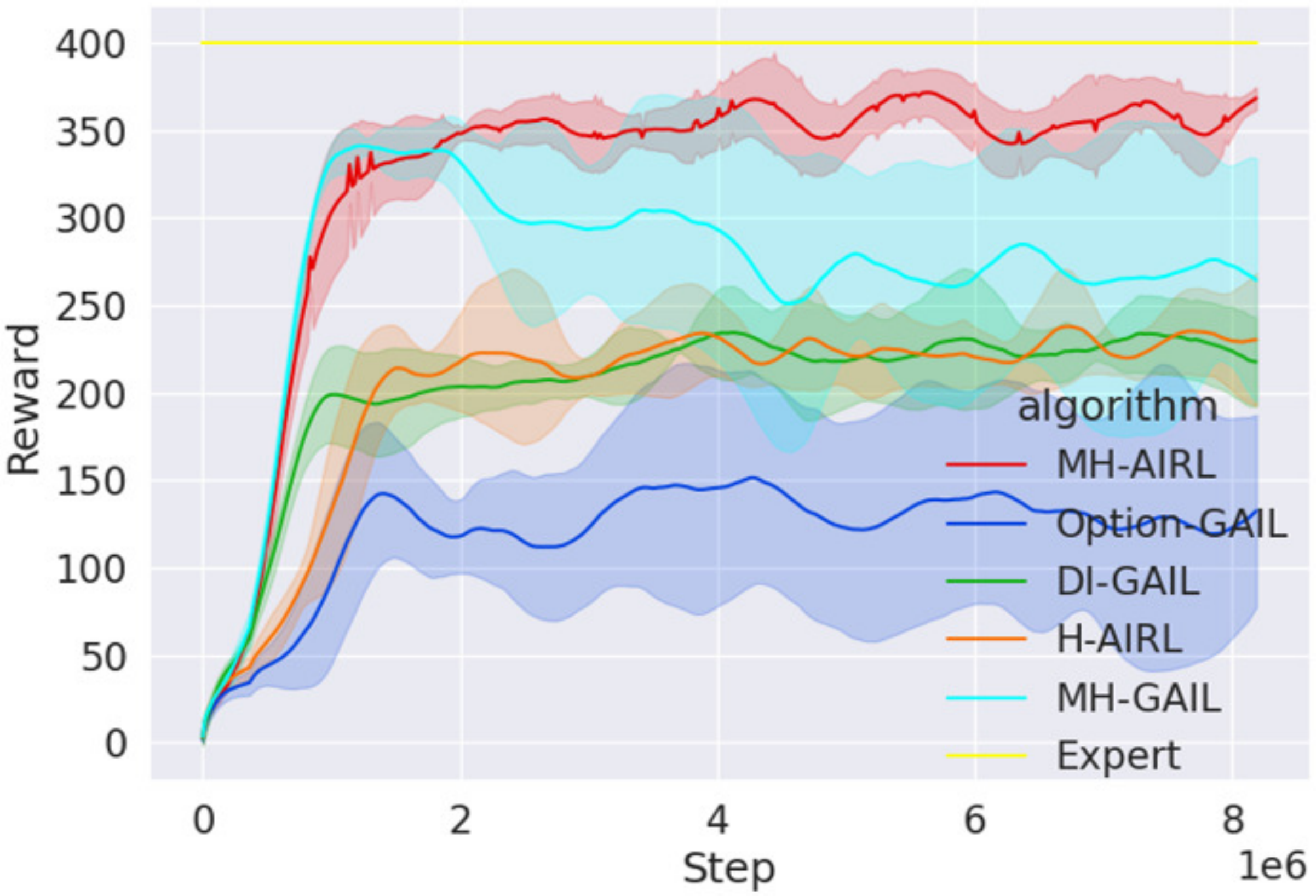}}
\subfigure[Ant-MultiGoal]{
\label{fig:5(c)} 
\includegraphics[width=2.4in, height=1.2in]{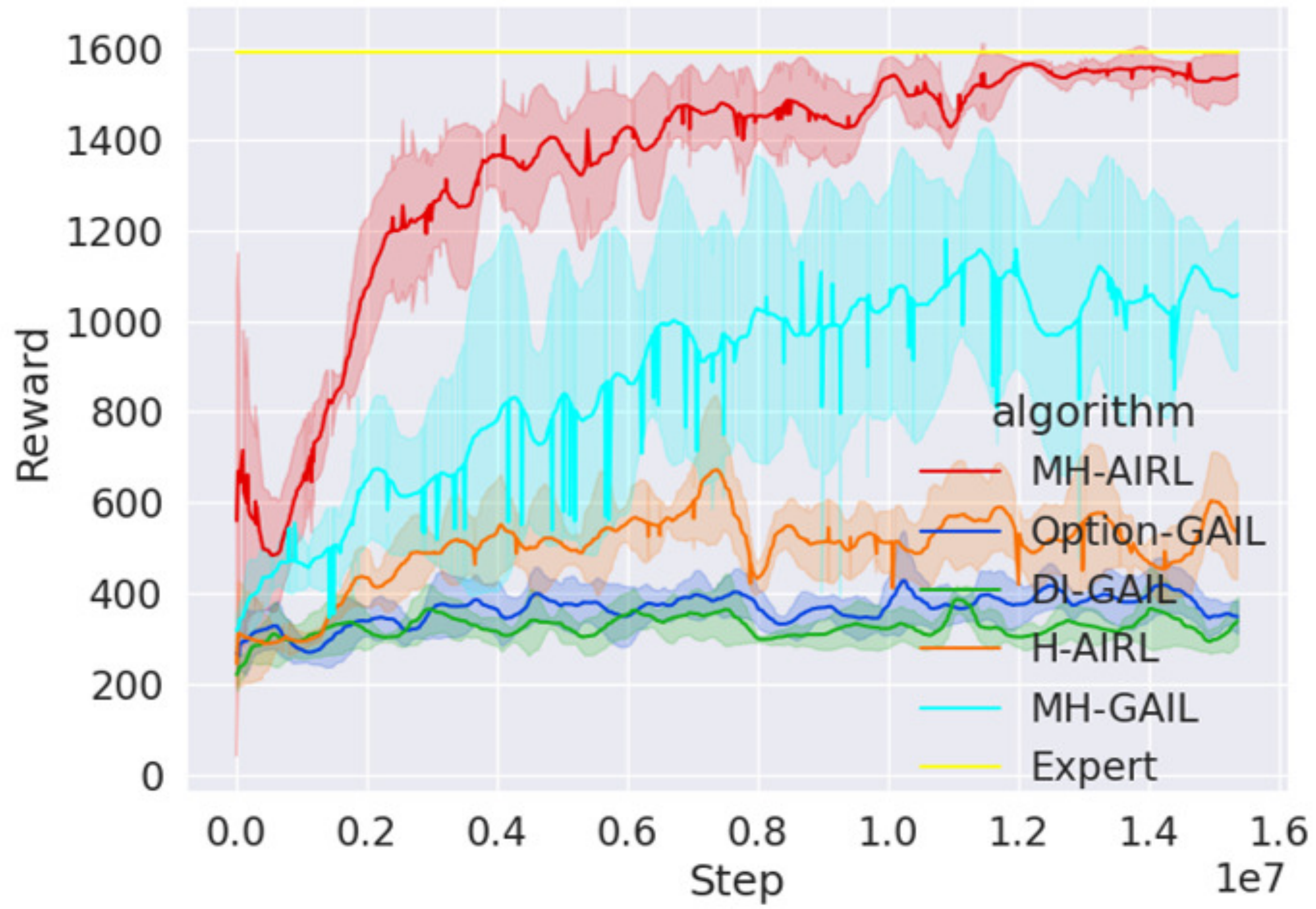}}
\subfigure[Kitchen-MultiSeq]{
\label{fig:5(d)} 
\includegraphics[width=2.4in, height=1.2in]{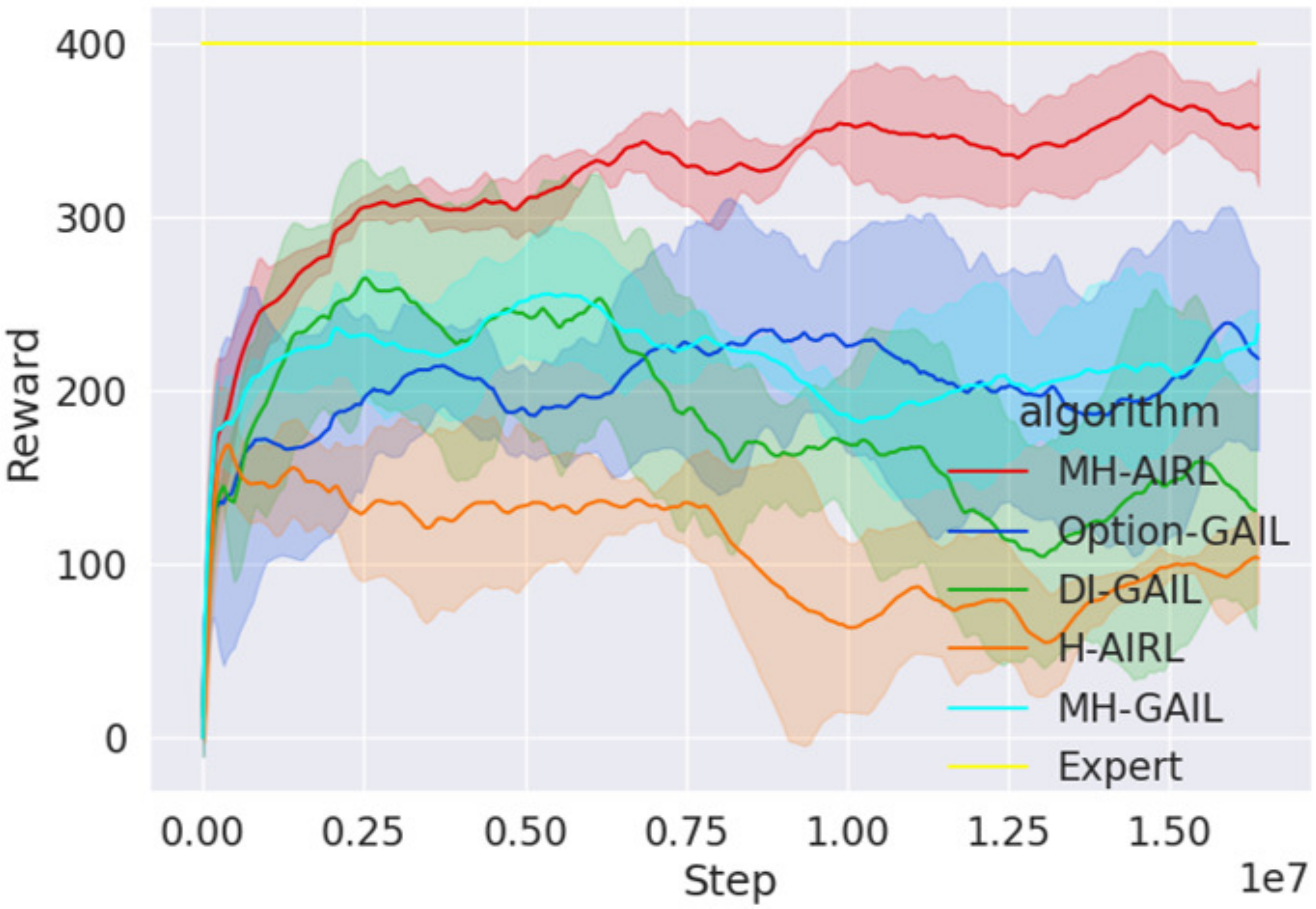}}
\caption{Comparison results of MH-AIRL with the ablated versions (MH-GAIL \& H-AIRL) and SOTA Hierarchical Imitation Learning (HIL) baselines (Option-GAIL \& DI-GAIL) on the four evaluation tasks. Our algorithm outperforms the baselines in all the tasks, especially in the more challenging ones (Ant \& Kitchen). MH-GAIL performs better than the other baselines which do not contain the multi-task learning component. H-AIRL, an ablation and HIL algorithm, has better performance than other SOTA HIL baselines on the Mujoco tasks. }
\label{fig:5} 
\end{figure*}

\subsection{Implementation Details of MH-GAIL} \label{mh-gail}

MH-GAIL is a variant of our algorithm by replacing the AIRL component with GAIL. Similar with Section \ref{h-airl}, we need to provide an extension of GAIL with the one-step option model, in order to learn a hierarchical policy. The extension method follows Option-GAIL \cite{DBLP:conf/icml/JingH0MKGL21} which is one of our baselines. MH-GAIL also uses an adversarial learning framework that contains a discriminator $D_{\vartheta}$ and a hierarchical policy $\pi_{\theta, \phi}$, for which the objectives are as follows:
\begin{equation} \label{equ:a350}
\begin{aligned}
        &\mathop{\max}_{\vartheta}\mathbb{E}_{C\sim prior(\cdot), (S, A, Z, Z') \sim \pi_{E}(\cdot|C)}\left[\log(1-D_{\vartheta}(S, A, Z, Z'| C))\right] + \\
        &\qquad\ \mathbb{E}_{C\sim prior(\cdot), (S, A, Z, Z') \sim \pi_{\theta, \phi}(\cdot|C)}\left[\log D_{\vartheta}(S, A, Z, Z'| C)\right] \\
        \mathop{\max}_{\theta, \phi}L^{IL}&=\mathop{\max}_{\theta, \phi}\mathbb{E}_{C\sim prior(\cdot), (X_{0:T}, Z_{0:T}) \sim \pi_{\theta, \phi}(\cdot|C)}\mathop{\sum}_{t=0}^{T-1}R_{IL}^t,\ R_{IL}^t = -\log D_{\vartheta}(S_t, A_t, Z_{t+1}, Z_t|C)
\end{aligned}
\end{equation}
where $(S, A, Z, Z')$ denotes $(S_t, A_t, Z_{t+1}, Z_t),\ t=\{0, \cdots, T-1\}$. It can be observed that the definition of $R_{IL}^t$ have changed. Moreover, the discriminator $D_{\vartheta}$ in MH-GAIL is trained as a binary classifier to distinguish the expert demonstrations (labeled as 0) and generated samples (labeled as 1), and does not have a specially-designed structure like the discriminator $D_{\vartheta}$ in MH-AIRL, which is defined with $f_{\vartheta}$ and $\pi_{\theta, \phi}$, so that it cannot recover the expert reward function.

\subsection{Analysis of the Learned Hierarchical Policy on HalfCheetah-MultiVel and Walker-RandParam} \label{halfwalker}

\begin{figure*}[htbp]
\centering
\subfigure[Results on HalfCheetah-MultiVel]{
\label{fig:7(a)} 
\includegraphics[width=4.0in, height=1.2in]{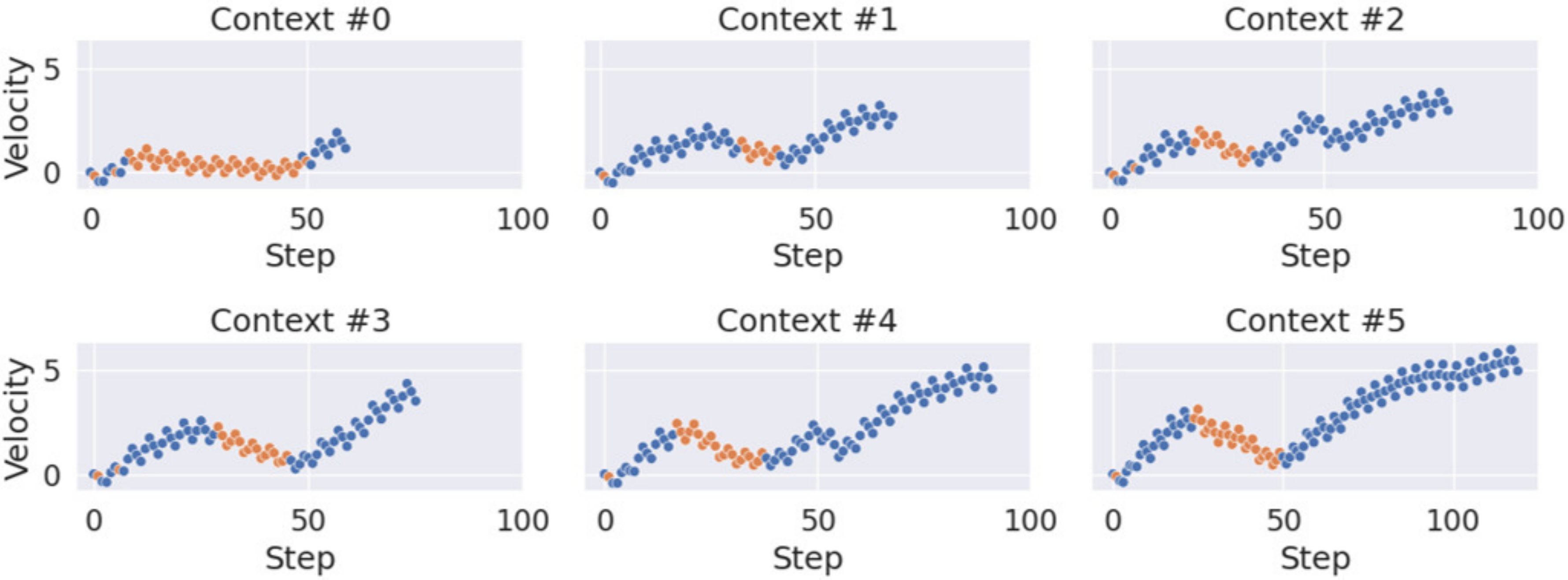}}
\subfigure[Results on Walker-RandParam]{
\label{fig:7(b)} 
\includegraphics[width=2.0in, height=1.2in]{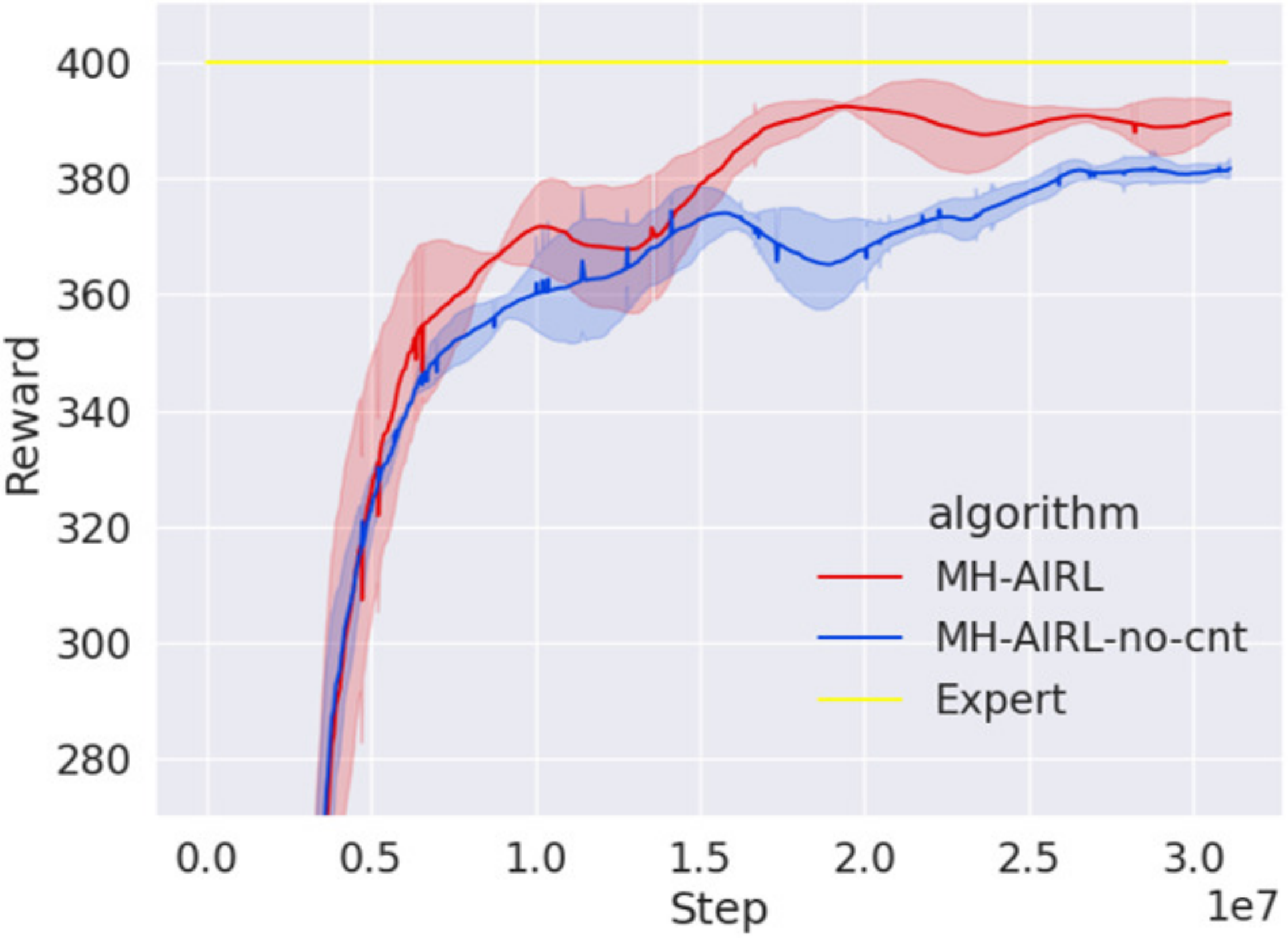}}
\caption{(a) Velocity change of the HalfCheetah agent in the test tasks with different goal velocities, where the agent adopts Option 1 (blue) and 2 (orange) when increasing and decreasing the speed, respectively. (b) For Walker-RandParam, the basic skills must adapt to the task setting, so the learning performance would drop without conditioning the low-level policy (i.e., option) on the task context.}
\label{fig:7} 
\end{figure*}

First, we randomly select 6 task contexts for HalfCheetah-MultiVel and visualize the recovered hierarchical policy as the velocity change of each episode in Figure \ref{fig:7(a)}. It can be observed that the agent automatically discovers two options (Option 1: blue, Option 2: orange) and adopts Option 1 for the acceleration phase ($0 \rightarrow v/2$ or $0 \rightarrow v$) and Option 2 for the deceleration phase ($v/2 \rightarrow 0$). This shows that MH-AIRL can capture the compositional structure within the tasks very well and transfer the learned basic skills to boost multi-task policy learning. 

Second, we note that, for some circumstances, the basic skills need to be conditioned on the task context. For the Mujoco-MultiGoal/MultiVel tasks, the basic skills (e.g., Option 2: decreasing the velocity) can be directly transferred among the tasks in the class and the agent only needs to adjust its high-level policy according to the task variable (e.g., adopting Option 2 when achieving $v/2$). However, for tasks like Walker-RandParam, the skills need to adapt to the tasks, since the mass of the agent changes and so do the control dynamics. As shown in Figure \ref{fig:7(b)},  the learning performance would drop without conditioning the low-level policy (i.e., option) on the task context, i.e., MH-AIRL-no-cnt.

\end{document}